\documentclass{article}

\PassOptionsToPackage{numbers, compress}{natbib}
    \usepackage[final]{neurips_2022}

\usepackage[utf8]{inputenc} % allow utf-8 input
\usepackage[T1]{fontenc}    % use 8-bit T1 fonts
\usepackage{hyperref}       % hyperlinks
\usepackage{url}            % simple URL typesetting
\usepackage{booktabs}       % professional-quality tables
\usepackage{amsfonts}       % blackboard math symbols
\usepackage{nicefrac}       % compact symbols for 1/2, etc.
\usepackage{microtype}      % microtypography
\usepackage{xcolor}         % colors
\usepackage{amsfonts}       % blackboard math symbols
\usepackage{subfigure} % 子图包
\usepackage{amsmath}
\usepackage{multirow}
\usepackage{color}
\usepackage{threeparttable} 
\usepackage{graphicx}
\newcommand{\red}[1]{\textcolor{red}{#1}}

\title{SCINet: Time Series Modeling and Forecasting with Sample Convolution and Interaction}

\author{Minhao Liu$^{*}$,  Ailing Zeng, Muxi Chen, Zhijian Xu, Qiuxia Lai, Lingna Ma, Qiang Xu$^{*}$ \\
\underline{CU}hk \underline{RE}liable Computing (CURE) Lab.\\
Dept. of Computer Science \& Egnineering, The Chinese University of Hong Kong\\
 $^{*}$\texttt{\{mhliu,qxu\}@cse.cuhk.edu.hk} \\
}

\begin{document}

\maketitle

\begin{abstract}
One unique property of time series is that the temporal relations are largely preserved after downsampling into two sub-sequences. By taking advantage of this property, we propose a novel neural network architecture that conducts sample convolution and interaction for temporal modeling and forecasting, named \textbf{SCINet}. Specifically, SCINet is a recursive \emph{downsample-convolve-interact} architecture. In each layer, we use multiple convolutional filters to extract \emph{distinct yet valuable} temporal features from the downsampled sub-sequences or features. By combining these rich features aggregated from multiple resolutions, SCINet effectively models time series with complex temporal dynamics. Experimental results show that SCINet achieves significant forecasting accuracy improvements over both existing convolutional models and Transformer-based solutions across various real-world time series forecasting datasets. Our codes and data are available at \url{https://github.com/cure-lab/SCINet}.
% \url{https://anonymous.4open.science/r/SCINet-2588}.
\end{abstract}

\section{Introduction}
Time series forecasting (TSF) enables decision-making with the estimated future evolution of metrics or events, thereby playing a crucial role in various scientific and engineering fields such as healthcare~\citep{Bahadori2019TemporalClusteringII}, energy management~\citep{Zhou2020InformerBE}, traffic flow~\citep{Zhou2020InformerBE}, and financial investment~\citep{DUrso2019TrimmedFC}, to name a few.

There are mainly three kinds of deep neural networks used for sequence modeling, and they are all applied for time series forecasting~\citep{Lim2021TimeseriesFW}: (i). recurrent neural networks (RNNs)~\citep{Hochreiter1997LongSM}; %and their variants such as long short-term memory (LSTM)~\citep{Hochreiter1997LongSM} and gated recurrent units (GRUs)~\citep{cho2014properties}; 
(ii). Transformer-based models~\citep{vaswani2017attention}; and (iii). temporal convolutional networks (TCN)~\citep{Bai2018AnEE}.

Despite the promising results of TSF methods based on these generic models, %applied in many real-world forecasting problems,
they do not consider the specialty of time series data during modeling.
For example, one unique property of time series is that the temporal relations (e.g., the trend and the seasonal components of the data) are largely preserved after downsampling into two sub-sequences. Consequently, by recursively downsampling the time series into sub-sequences, we could obtain a rich set of convolutional filters to extract dynamic temporal features at multiple resolutions. 
Motivated by the above, in this paper, we propose a novel neural network architecture for time series modeling and forecasting, named \textit{sample convolution and interaction network} (\textit{SCINet}). The main contributions of this paper are as follows:  %Other types of sequence data do not have such characteristics.  
%Existing TSF works, however, simply apply generic sequence models without taking the unique properties of time series into consideration. 

\begin{figure*}[t]
    \centering
    \includegraphics[width=0.95\textwidth]{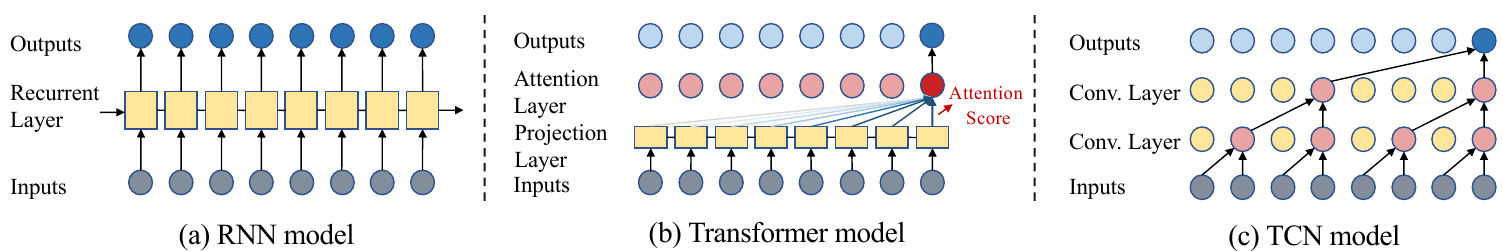}
    \caption{Existing sequence modeling architectures for time series forecasting.}
    \label{fig:related}
     \vspace{-5pt}
\end{figure*}

\begin{itemize}
%\vspace{-5pt}
%\item We discover the misconception of existing TCN design principles for the TSF problem. In particular, we show causal convolution is \textbf{NOT} necessary and better prediction accuracy can be achieved by removing such constraints. 

\item We propose SCINet, a hierarchical \emph{downsample-convolve-interact} TSF framework that effectively models time series with complex temporal dynamics. By iteratively extracting and exchanging information at multiple temporal resolutions, an effective representation with enhanced predictability can be learned, as verified by its comparatively lower permutation entropy (PE)~\citep{Huang2019EnhancedTS}.
\vspace{5pt}
\item We design the basic building block, \textit{SCI-Block}, for constructing SCINet, which downsamples the input data/feature into two sub-sequences, and then extracts features of each sub-sequence using distinct convolutional filters. To compensate for the information loss during the downsampling procedure, we incorporate interactive learning between the two convolutional features within each SCI-Block.
\end{itemize}

Extensive experiments on various real-world TSF datasets show that our model consistently outperforms existing TSF approaches by a considerable margin. Moreover, while SCINet does not explicitly model spatial relations, it achieves competitive forecasting accuracy on spatial-temporal TSF tasks.   %in~\cite{Zhou2020InformerBE}, on average, SCINet achieves more than $40\%$ relative improvement compared to state-of-the-art methods, in terms of mean squared errors of the prediction results. }

%\vspace{-10pt}
\section{Related Work and Motivation}
\label{sec:related}
The time series forecasting problem is defined as: Given a long time series $\mathbf{X}^*$ and a look-back window of fixed length $T$, at timestamp $t$, 
%single-step forecasting is to predict the future value $\hat{\mathbf{X}}_{t+\tau:t+\tau} = \{\mathbf{x}_{t+ \tau}\}$, while multi-step forecasting is to predict multiple future values  
time series forecasting is to predict $\hat{\mathbf{X}}_{t+1:t+\tau} = \{ \mathbf{x}_{t+1},..., \mathbf{x}_{t+ \tau} \}$ based on the past $T$ steps $\mathbf{X}_{t-T+1:t} = \{ \mathbf{x}_{t-T+1},..., \mathbf{x}_{t} \}$. Here, $\tau$ is the length of the forecast horizon, $\mathbf{x}_{t} \in \mathbb{R}^{d}$ is the value at time step $t$, and $d$ is the number of variates. 
For simplicity, in the following we will omit the subscripts, and use $\mathbf{X}$ and $\hat{\mathbf{X}}$ to represent the historical data and the forecasted data, respectively.
%When $\tau>1$, we can either directly optimize the multi-step forecasting objective (\emph{direct multi-step (DMS)} estimation), or learn a single-step forecaster and iteratively apply it to get multi-step predictions (\emph{iterated multi-step (IMS)} estimation)~\citep{shi2018machine}.

\subsection{Related Work}

Traditional time series forecasting methods such as the autoregressive integrated moving average (ARIMA) model~\citep{Box1968SomeRA} and Holt-Winters seasonal method~\citep{Holt2004ForecastingSA} have theoretical guarantees. However, they are mainly applicable for univariate forecasting problems, restricting their applications to complex time series data. With the increasing data availability and computing power in recent years, it is shown that deep learning-based TSF techniques have the potential to achieve better forecasting accuracy than conventional approaches~\citep{Lim2021TimeseriesFW,Oreshkin2020NBeats}. 

%In this section, we focus on deep learning-based methods for TSF due to their superior performance, especially for multivariate time series ($d>1$). 
%\subsection{Related Work}
%There are mainly three kinds of deep neural networks for TSF.
Earlier RNN-based TSF methods~\citep{rangapuram2018deep,salinas2020deepar} summarize the past information compactly in the internal memory states that are recursively updated with new inputs at each time step, as shown in Fig.~\ref{fig:related}(a). 
%These methods generally belong to IMS estimation methods, 
% which obtain the multi-step predictions by recursively feeding previous one-step prediction into future inputs, 
%which suffer from error accumulation. 
The gradient vanishing/exploding problems and the inefficient training procedure greatly restrict the application of RNN-based models. 

In recent years, Transformer-based models~\citep{vaswani2017attention} have taken the place of RNN models in almost all sequence modeling tasks, thanks to the effectiveness and efficiency of the self-attention mechanisms. Various Transformer-based TSF methods (see Fig.~\ref{fig:related}(b)) are proposed in the literature~\citep{li2019enhancing,lim2021temporal,Wu2021AutoformerDT,Liu2022Pyraformer}. These works typically focus on the challenging long-term time series forecasting problem, taking advantage of their remarkable long sequence modeling capabilities.
%, as discussed in Section~\ref{sec:rethinking_transformer}. 
%, and they are shown to be quite effective in predicting long sequences. 
%However, the overhead of Transformer-based models is a serious concern and lots of research efforts are dedicated to tackling this problem (e.g.,~\cite{Zhou2020InformerBE}).

Another popular type of TSF model is the so-called temporal convolutional network~\cite{borovykh2017conditional,Bai2018AnEE,Sen2019ThinkGA, Wu2019GraphWF,nguyen2021temporal}, wherein convolutional filters are used to capture local temporal features (see Fig.~\ref{fig:related}(c)). The proposed SCINet is also constructed based on temporal convolution. However, our method has several key differences compared with the TCN model based on dilated causal convolution, as discussed in the following.

\subsection{Rethinking Dilated Causal Convolution for Time Series Modeling and Forecasting}
\label{sec:rethinking_tcn}

The local correlation of time series data is reflected in the continuous changes within a time slot, and convolutional filters can effectively capture such local features. Consequently, convolutional neural networks are explored in the literature for time series modeling and forecasting. In particular, dilated causal convolution (DCS) is the current \emph{de facto} method used in this respect. 
%One of the critical challenges in using convolutional network for time series modeling is how to expand the receptive field for long sequences. 

DCS was first proposed for generating raw audio waveforms in WaveNet~\citep{Oord2016WaveNetAG}. Later,~\citep{Bai2018AnEE} simplifies the WaveNet architecture to the so-called temporal convolutional networks (see Fig.~\ref{fig:related} (c)). TCN consists of a stack of causal convolutional layers with exponentially enlarged dilation factors, which can achieve a large receptive field with just a few convolutional layers. Over the years, TCN has been widely used in all kinds of time series forecasting problems and achieve promising results~\citep{Wu2019GraphWF,Sen2019ThinkGA}. Moreover, convolutional filters can work seamlessly with graph neural networks (GNNs) to solve various spatial-temporal TSF problems. 

With \emph{causal convolutions} in the TCN architecture, an output $i$ is convolved only with the $i^{th}$ and earlier elements in the previous layer. While causality should be kept in forecasting tasks, the potential ``future information leakage" problem exists only when the output and the input have temporal overlaps. In other words, causal convolutions should be applied only in autoregressive forecasting, wherein the previous output serves as the input for future prediction. When the predictions are completely based on the known inputs in the look-back window, there is no need to use causal convolutions. We can safely apply \textit{normal convolutions} on the look-back window for forecasting. 

More importantly, the dilated architecture in TCN has two inherent limitations:

\begin{itemize}
    \item A single convolutional filter is shared within each layer. Such a unified convolutional kernel tends to extract the average temporal features from the data/features in the previous layer. However, complex time series may contain substantial temporal dynamics. Hence, it is essential to extract distinct yet valuable features with a rich set of convolutional filters.
    \item  While the final layer of the TCN model has the global view of the entire look-back window, the effective receptive fields of the intermediate layers (especially those close to the inputs) are limited, causing temporal relation loss during feature extraction.  
\end{itemize}

The above limitations of the TCN architecture motivate the proposed SCINet design, as detailed in the following section. 
%restricting the extraction capability of complex temporal patterns. 
%significantly restricting the extraction of temporal dynamics from the data/features in the previous layer. 
%\red{More importantly, while the dilation procedure enables global view of on the look-back window, it does not extract temporal information at multiple gra ... }

%In view of the relatively high accuracy of TCN in many forecasting tasks despite the above limitations, we propose a novel downsample-convolve-interact architecture \textit{SCINet} that takes the unique properties of time series into consideration to achieve better prediction accuracy, as detailed in the following section. 

\begin{figure*}[htbp]	
\centering
\includegraphics[width=0.95\textwidth]{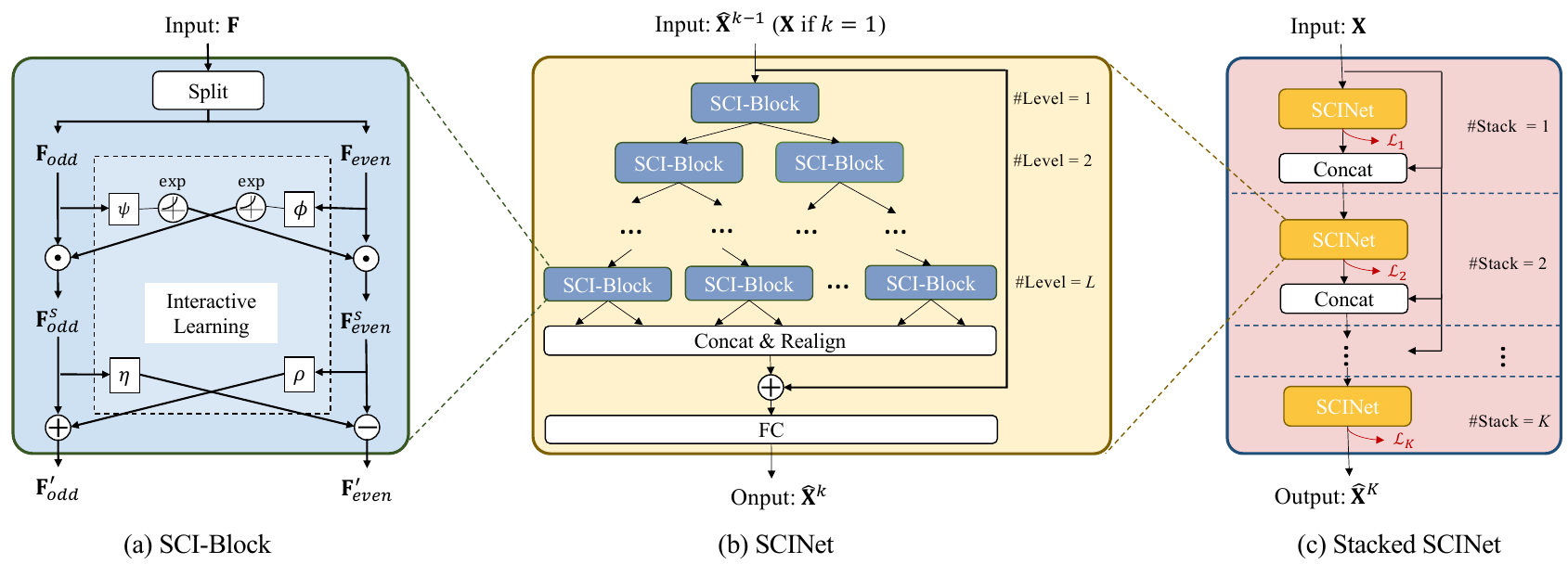}
% \vspace{-5pt}
\caption{The overall architecture of Sample Convolution and Interaction Network (SCINet).}
\label{fig:frame}
\vspace{-5pt}
\end{figure*}

\section{SCINet: Sample Convolution and Interaction Network}
\label{sec:sci_network}

%The rationale behind the proposed SCINet design is that complex time series contain substantial temporal dynamics, and it is essential to capture such features with a rich set of convolutional filters that have both local and global views of the time series. 
%The proposed \textit{Sample Convolution and Interaction Network} (\textit{SCINet}) is hence designed as hierarchical framework that enhances the predictability of the original time series by capturing temporal dependencies at multiple temporal resolutions\footnote{\textit{Multi-resolution analysis} is a classic method for signal analysis~\citep{mallat1989theory} and has been used in time series classification tasks~\citep{Cui2016MultiScaleCN}.}. 

SCINet adopts an encoder-decoder architecture. The encoder is a hierarchical convolutional network that captures dynamic temporal dependencies at multiple resolutions with a rich set of convolutional filters. As shown in Fig.~\ref{fig:frame}(a), the basic building block, \textit{SCI-Block} (Section~\ref{sec:sciblock}), downsamples the input data or feature into two sub-sequences and then processes each sub-sequence with a set of convolutional filters to extract distinct yet valuable temporal features from each part. To compensate for the information loss during downsampling, we incorporate \textit{interactive learning} between the two sub-sequences. 
Our \textit{SCINet} (Section~\ref{sec:scinet}) is constructed by arranging multiple SCI-Blocks into a binary tree structure (Fig.~\ref{fig:frame}(b)). A distinctive advantage of such design is that each SCI-Block has both local and global views of the entire time series, thereby facilitating the extraction of useful temporal features. 
After all the downsample-convolve-interact operations, we realign the extracted features into a new sequence representation and add it to the original time series for forecasting with a fully-connected network as the decoder.  
%To allow accumulation of the historical information when the length of the input sequence is relatively short, 
To facilitate extracting complicated temporal patterns, we could further stack multiple SCINets and apply intermediate supervision to get a \textit{Stacked SCINet} (Section~\ref{sec:stacked_scinet}), as shown in Fig.~\ref{fig:frame}(c).

% \begin{table*}[t]	
% \caption{The datasets used in our experiments.}
% \vspace{-5pt}
% \begin{center}
% % \footnotesize
% % \resizebox{\textwidth}{!}
% % {
% \begin{tabular}{c|ccc|cccc} 
% \hline
% Datasets       & ETTh1      & ETTh2      & ETTm1      & PEMS03       & PEMS04      & PEMS07      & PEMS08      \\ \hline
% \# Variates    & 7          & 7          & 7                       & 358          & 307         & 883         & 170         \\
% \# Timesteps   & 17,420     & 17,420     & 69,680              & 26,209       & 16,992      & 28,224      & 17,856      \\
% Granularity    & 1hour      & 1hour      & 15min                & 5min         & 5min        & 5min        & 5min        \\
% Start time     & 7/1/2016   & 7/1/2016   & 7/1/2016         & 5/1/2012     & 7/1/2017    & 5/1/2017    & 3/1/2012    \\
% Task type      & Multi-step & Multi-step & Multi-step   & Multi-step   & Multi-step  & Multi-step  & Multi-step  \\ \hline
% Data partition & \multicolumn{3}{c|}{Training/Validation/Testing: 12/4/4 months}           & \multicolumn{4}{c}{Training/Validation/Testing: 6/2/2 split ratio} \\ \hline
% \end{tabular}
% \end{center}

% \label{tab:datasets}
% \vspace{-5pt}
% \end{table*}

\subsection{SCI-Block}\label{sec:sciblock}

% downsample

The SCI-Block (Fig.~\ref{fig:frame}(a)) is the basic module of the SCINet, which decomposes the input feature $\mathbf{F}$ into two sub-features $\mathbf{F}^{'}_{odd}$ and $\mathbf{F}^{'}_{even}$ through the operations of \emph{Spliting} and \emph{Interactive-learning}. 
%Please note, our SCI-Block design requires no domain knowledge %is required to manually design the downsampling rates. Thus 
%and it can be easily generalized to various time series data. 

The \textit{Splitting} procedure downsamples the original sequence $\mathbf{F}$ into two sub-sequences $\mathbf{F}_{even}$ and $\mathbf{F}_{odd}$ by separating the even and the odd elements, which are of coarser temporal resolution but preserve most information of the original sequence.

Next, we use different convolutional kernels to extract features from $\mathbf{F}_{even}$ and $\mathbf{F}_{odd}$. As the kernels are separate, the extracted features from them would contain distinct yet valuable temporal relations with enhanced representation capabilities.
% Considering the heterogeneity information of each downsampled sub-sequence, we propose to process $\mathbf{F}_{even}$ and $\mathbf{F}_{odd}$ using two distinct sets of convolution layers.
% \textbf{Interactive learning.} 
To compensate for potential information loss with downsampling, 
%after extracting features from $\mathbf{F}_{even}$ and $\mathbf{F}_{odd}$ using distinct convolution layers, 
we propose a novel \emph{interactive-learning} strategy to allow information interchange between the two sub-sequences by learning affine transformation parameters from each other. 
% is based on the strong correlation among the neighbouring samples to extract the implicit temporal dependencies, 
As shown in Fig.~\ref{fig:frame} (a), the interactive learning procedure consists of two steps. 
% which can greatly enhance the representation ability of the learned features by coupling the information of the two complementary parts.

First, $\mathbf{F}_{even}$ and $\mathbf{F}_{odd}$ are projected to hidden states with two different 1D convolutional modules $\phi$ and $\psi$, respectively, and transformed to the formats of $\exp$ and interact to the $\mathbf{F}_{even}$ and $\mathbf{F}_{odd}$ with the element-wise product (see Eq.~(\ref{eq:inter_1})). 
This can be viewed as performing scaling transformation on $\mathbf{F}_{even}$ and $\mathbf{F}_{odd}$, where the scaling factors are learned from each other using neural network modules.
% \red{This step makes the local information from each sub-feature diffuse to the global to ensure the valuable information interchange in a global perspective in the next step. The $\exp$ format is better for model training.} 
Here, $\odot$ is the Hadamard product or element-wise production.

\vspace{-10pt}
\begin{equation}\label{eq:inter_1}
\small
    \mathbf{F}^{s}_{odd} = \mathbf{F}_{odd}\odot   \exp(\phi (\mathbf{F}_{even}) ),~~~
    \mathbf{F}^{s}_{even} = \mathbf{F}_{even}\odot \exp(\psi (\mathbf{F}_{odd}) ).
\end{equation}
\vspace{-10pt}
\begin{equation}\label{eq:inter_2}
\small
    \mathbf{F}^{'}_{odd} = \mathbf{F}^{s}_{odd} \pm  \rho( \mathbf{F}^{s}_{even}),~~~
    \mathbf{F}^{'}_{even} = \mathbf{F}^{s}_{even} \pm   \eta(\mathbf{F}^{s}_{odd}).
\end{equation}
%\vspace{2pt}

Second, as shown in Eq.~(\ref{eq:inter_2}), the two scaled features $\mathbf{F}^{s}_{even}$ and $\mathbf{F}^{s}_{odd}$ are further projected to another two hidden states with the other two 1D convolutional modules $\rho$ and $\eta$, and then added to or subtracted from\footnote{The selection of the operators in Eq.(2) affects the parameter initialization of our model and we show its impact in the Appendix~\ref{appendix:B3}.} $\mathbf{F}^{s}_{even}$ and $\mathbf{F}^{s}_{odd}$.
% the "addition" summarizes the overall information from the two sub-sequences, while the "subtraction" emphasizes the difference between them. Intuitively, such a design makes it easier for the network to aggregate useful information according to the given supervision, see Appendix.}). 
The final outputs of the interactive learning module are two updated sub-features $\mathbf{F}^{'}_{even}$ and $\mathbf{F}^{'}_{odd}$. The default architectures of $\phi$, $\psi$, $\rho$ and $\eta$ are shown in the Appendix ~\ref{appendix:reprod}. 

Compared to the dilated convolutions used in the TCN architecture, the proposed downsample-convolve-interact architecture achieves an even larger receptive field at each convolutional layer. More importantly, unlike TCN that employs a single shared convolutional filter at each layer, significantly restricting its feature extraction capabilities, SCI-Block aggregates essential information extracted from the two downsampled sub-sequences that have both local and global views of the entire time series.% according to the given supervision for prediction.

%\red{To avoid the information loss, we keep both sub-sequences for feature extraction. This is different from previous practice in multi-scale analysis where only a single sub-sequence at a certain temporal resolution is used for feature extraction. }

\subsection{SCINet}
\label{sec:scinet}

% Our \textit{SCINet} (Section~\ref{sec:scinet}) is constructed by arranging multiple SCI-Blocks into a binary tree structure (Fig.~\ref{fig:frame} (b)). 
% After all the downsample-convolve-interact operations, we realign and concatenate all the low-resolution components into a new sequence representation, and add it to the original time-series for forecasting. 
% \red{level yue da , di ceng de jiu bao le geng duo de multi-resolution de xin xi}
With the SCI-Blocks presented above, we construct the SCINet by arranging multiple SCI-Blocks hierarchically and get a tree-structured framework, as shown in Fig.~\ref{fig:frame} (b).  
% The SCINet is formed as a hierarchical tree structure, and the nodes of the tree are a series of SCI-Blocks. 

There are $2^{l}$ SCI-Blocks at the $l$-th level, where $l= 1, \dots, L$ is the index of the level, and $L$ is the total number of levels. 
% For the $1$st SCINet in the stacked SCINet, the input is $\mathbf{X}$.
Within the $k$-th SCINet of the stacked SCINet~(Section~\ref{sec:stacked_scinet}), the input time series $\mathbf{X}$ (for $k\!=\!1$) or feature vector $\hat{\mathbf{X}}^{k-1}\!=\!\{\hat{\mathbf{x}}_{1}^{k-1}, ..., \hat{\mathbf{x}}_{\tau}^{k-1}\}$ (for $k\!>\!1$) is gradually down-sampled and processed by SCI-Blocks through different levels, which allows for effective feature learning of different temporal resolutions. 
%This is because the temporal resolution of the features gradually decreases when $l$ increases.
% This is because the SCI-Blocks at different levels are responsible for features of different temporal resolutions, i.e., when $l$ increases, the temporal resolution decreases.  
% \red{and the features of the deeper levels contain more temporal resolutions information transmitted from the shallow levels.} 
In particular, the information from previous levels will be gradually accumulated, i.e., the features of the deeper levels would contain extra finer-scale temporal information transmitted from the shallower levels. 
In this way, we can capture both short-term and long-term temporal dependencies in the time series. 

After going through $L$ levels of SCI-Blocks, we rearrange the elements in all the sub-features by reversing the odd-even splitting operation and concatenate them into a new sequence representation. It is then added to the original time series through a residual connection~\citep{He2016DeepRL} to generate a new sequence with enhanced predictability. 
Finally, a simple fully-connected network is used to decode the enhanced sequence representation into $\hat{\mathbf{X}}^{k}\!=\!\{\hat{\mathbf{x}}_{1}^{k}, ..., \hat{\mathbf{x}}_{\tau}^{k}\}$. %which represents a fused feature or the final prediction depending on the location of current SCINet in the stacked SCINet. 

% This fully-connected layer also enables an output of a different l ength with the input, as opposed to the TCN where the length of the output is forced to be equal to that of the input. 

\iffalse
The number of levels $L$ mainly depends on the size of the look-back window $T$. 
Although increasing $L$ would introduce extra computational cost, we will show that SCI-Net is still more efficient than other convolution-based methods such as TCN. 
As shown in Fig.~\ref{fig:related} (c), TCN employs convolutions with exponentially enlarged dilation factors. 
The number of convolution layers needed for TCN to cover a look-back window of length $T$ is $\log_2T$, and the resulted number of convolution operations is $T*\log_2T$. 
On the contrary, SCINet does not rely on dilation to enlarge the effective receptive fields of feature extraction, thanks to interactive learning which allows global information interchange between the two downsampled parts. Thus, we do not need $\log_2T$ levels to cover the look-back window, and the number of convolution operations of SCINet is $L*2T$, which can be much less than that of the TCN. As experimentally verified in Section~\ref{sec:ab},  $L\!\leq\!5$ would satisfy the requirements of most cases.
\fi

\subsection{Stacked SCINet}
\label{sec:stacked_scinet}

When there are sufficient training samples, we could stack $K$ layers of SCINets to achieve even better forecasting accuracy (see Fig.~\ref{fig:frame} (c)), at the cost of a more complex model structure. 

\iffalse
When the length of the look-back window $T$ is comparable with the length of the prediction horizon $\tau$, it would be difficult for a single SCINet to fully capture the temporal dependencies. 
\red{To keep the continuous temporal correlations between the look-back window and the target prediction horizon, meanwhile fully accumulate the historical information within the look-back window, we can further stack $K$ SCINets end-to-end to form the \textit{Stacked SCINet} structure,
% where the output of one SCINet is feeded as the input into the next, 
as illustrated in Fig.~\ref{fig:frame} (c). } 
\fi

Specifically, we apply \textit{intermediate supervision}~\citep{bai2018trellis} on the output of each SCINet using the ground-truth values, to ease the learning of the intermediate temporal features.
% \red{Besides, to model the continuous temporal correlations between the lookback window and the targe prediction horizon in later stacks, we choose to concatenate part of the original input sequence.}
The output of the $k$-th intermediate SCINet, $\hat{\mathbf{X}}^{k}$ with length $\tau$, is concatenated with part of the input $\mathbf{X}_{t-(T-\tau)+1:t}$ to recover the length to the original input and feeded as input into the $(k\!+\!1)$-th SCINet, where $k=1, \dots, K\!-\!1$, and $K$ is the total number of the SCINets in the stacked structure.
% In this way, the long-term correlations between input and prediction are utilized to aid the learning of temporal dependencies, which is similar to TrellisNet~\citep{Bai2019TrellisNF} that have demonstrated strong performance with multiple iterative stages and intermediate supervision. 
The output of the $K$-th SCINet, $\hat{\mathbf{X}}^{K}$, is the final forecasting results. 
%Similar to the analysis on $L$, increasing $K$ would also lead to higher computational cost.

\subsection{Loss Function}
% \textbf{Single-step forecasting.}
To train a stacked SCINet with $K$ ($K\geq 1$) SCINets, the loss of the $k$-th prediction results is calculated as the L1 loss between the output of the $k$-th SCINet and the ground-truth horizontal window to be predicted:
\begin{equation} \label{eq:mid_loss}
            % \mathcal{L}_{k} = \red{\frac{1}{\tau}} \left \| \hat{\mathbf{X}}^{k+1}  - \hat{\mathbf{X}}  \right \|, ~~~1\leq k<K .
            \mathcal{L}_{k} =  \frac{1}{\tau}\sum_{i=0}^{\tau}\left \| \hat{\mathbf{x}}^{k}_{i} - \mathbf{x}_{i} \right \| %,~~~k\neq K .
\end{equation}
% where $\mathbf{x}^{k+1}$ is the output of the $k$-th SCINet, and $\hat{\mathbf{x}}$ is the horizontal window to be predicted.

% The loss function of the last SCINet depends on the type of forecasting tasks. For multi-step forecasting, the loss $\mathcal{L}_{K}$ is the same as Eq.~(\ref{eq:mid_loss}). For single-step forecasting, we introduce a balancing parameter $\lambda \in (0,1)$ for the value of the last time-step\footnote{This is slightly different from other practice for single-step forecasting~\citep{Lai2018ModelingLA}, because we choose to use all the available values in the prediction window as supervision signal.}, which will be discussed in Appendix:
% \begin{equation}\label{eq:last_loss}
%              \mathcal{L}_{K} = \frac{1}{\tau-1}\sum_{i=0}^{\tau-1}\left \| \hat{\mathbf{x}}^{K}_{i} - \mathbf{x}_{i} \right \| + \lambda \left \| \hat{\mathbf{x}}^{K}_{\tau} - \mathbf{x}_{\tau} \right \|.
% \end{equation}

The total loss of the stacked SCINet can be written as:
\begin{equation}
     \mathcal{L} = \sum_{k=1}^{K}\mathcal{L}_{k}.
\end{equation}

\subsection{Complexity Analysis}
Thanks to the downsampling procedure, the neurons at each convolutional layer of SCINet have a larger receptive field than those of TCN. More importantly, the set of rich convolutional filters in SCINet enable flexible extraction of temporal features from multiple resolutions. Consequently, SCINet usually does not require downsampling the original sequence to the coarsest level for effective forecasting. Given the look-back window size $T$, TCN generally requires $\lceil\log_2T\rceil$ layers when the dilation factor is 2, while the number of layers $L$ in SCINet could be much smaller than $\log_2T$. Our empirical study shows that the best forecasting accuracy is achieved with $L\!\leq\!5$ in most cases even with large $T$ (e.g., $168$). As for the number of stacks $K$, our empirical study also shows that $K\!\leq\!3$ would be sufficient.

Consequently, the computational cost of SCINet is usually on par with that of the TCN architecture. The worst-case time complexity is $\mathcal{O}(T\log T)$, much less than that of vanilla Transformer-based solutions: $\mathcal{O}(T^2)$.

%The time complexity of the SCINet depends on the number of levels $l$, and the maximum value of $l$ is $log_2T$ ($T$ is the size of lookback window). Therefore, the \textbf{worst-case} time complexity of SCINet is $\mathcal{O}(T\log T)$ which is the same as that of TCN: $\mathcal{O}(T\log T)$. Besides, it is much less than that of vanilla Transformer-based solutions: $\mathcal{O}(T^2)$.

\iffalse
\subsection{Overlap-FC}

\red{Instead of using single layer of simple fully connected layer as decoder, we designed an over-lapped fully connected layer as our decoder, as shown in fig.xx. Comparing with the original FC layer, our over-lapped FC layer can bring xx\% of performance boost.
}
\fi

\section{Experiments}
\label{sec:exp}

In this section, we show the quantitative and qualitative comparisons with the state-of-the-art models for time series forecasting. We also present a comprehensive ablation study to evaluate the effectiveness of different components in SCINet. More details on \emph{datasets}, \emph{evaluation metrics}, \emph{data pre-processing}, \emph{experimental settings}, \emph{network structures} and their \emph{hyper-parameters} are shown in the Appendix.

\iffalse
In Section~\ref{sec:datasets}, we first briefly introduce the $7$ datasets on which we train the model and make comparisons with other models. 
In Section~\ref{sec:comparison_sota}, we show the quantitative and qualitative comparisons with the state-of-the-arts, and analyze the predictability of our method. 
% Then, we perform predictability analysis on SCINet in Section~\ref{sec:interpretation}. 
% try to explain why SCINet achieves better forecasting performances by measuring the predictability of the time series in Section~\ref{sec:interpretation}. 
% Then, to further demonstrate the interpretation of our approach, we visualize the results before and after the residual manipulation and utilize the permutation entropy to evaluate the intrinsic predictability of the sequence.
Finally, a comprehensive ablation study is conducted in Section~\ref{sec:ab} to assess the effectiveness of different components. More details on \emph{datasets}, \emph{evaluation metrics}, \emph{data pre-processing}, \emph{experimental settings} and \emph{network structure $\&$ hyper-parameter tuning} are shown in the Appendix. % A and B.
\fi

\subsection{Datasets}
\label{sec:datasets}
% \textbf{Dataset}
% 把数据集分下类,不放etth2和m1的话要删table 1中对应的

We conduct experiments on $11$ popular time series datasets: (1)~\emph{Electricity Transformer Temperature}~\citep{Zhou2020InformerBE}~(ETTh) 
(2)~\emph{Traffic}
(3)~\emph{Solar-Energy}
(4)~\emph{Electricity}
(5)~\emph{Exchange-Rate}
(6)~\textit{PeMS} (\emph{PEMS03, PEMS04, PEMS07 and PEMS08}). 
A brief description of these datasets is listed in Table~\ref{tab:datasets}. All the experiments on these datasets in this section are conducted under multi-variate TSF setting. 

To make a fair comparison, we follow existing experimental settings, and use the same evaluation metrics as the original publications \citep{Hyndman2006AnotherLA, Makridakis1982TheAO, xu2021autoformer, Lai2018ModelingLA} in each dataset. 

\begin{table*}[t]
\caption{The overall information of the $11$ datasets.}
% \tiny
\begin{center}
% \footnotesize
\scriptsize
\resizebox{\textwidth}{!}
{
\begin{tabular}{c|cc|cccc|cccc}
\hline
Datasets       & ETTh(1,2)     & ETTm1          & Traffic     & Solar-Energy & Electricity & Exchange-Rate & PEMS03       & PEMS04      & PEMS07      & PEMS08      \\ \hline
Variants       & 7             & 7              & 862         & 137          & 321         & 8             & 358          & 307         & 883         & 170         \\
Timesteps      & 17,420        & 69,680         & 17,544      & 52,560       & 26,304      & 7,588         & 26,209       & 16,992      & 28,224      & 17,856      \\
Granularity    & 1hour         & 15min          & 1hour       & 10min        & 1hour       & 1day          & 5min         & 5min        & 5min        & 5min        \\
Start time     & 7/1/2016      & 7/1/2016       & 1/1/2015       & 1/1/2006    & 1/1/2012     & 1/1/1990    & 5/1/2012      & 7/1/2017     & 5/1/2017    & 3/1/2012    \\
Task type      & Multi-step    & Multi-step     & Single-step & Single-step  & Single-step & Single-step   & Multi-step   & Multi-step  & Multi-step  & Multi-step  \\ \hline
Data partition & \multicolumn{2}{c|}{Follow~\citep{Zhou2020InformerBE}}          & \multicolumn{4}{c|}{Training/Validation/Testing: 6/2/2}  & \multicolumn{4}{c}{Training/Validation/Testing: 6/2/2} \\ \hline
\end{tabular}}
\end{center}

\label{tab:datasets}
\end{table*}

% To make a fair comparison, we follow some popular experiment setting, and use the same evaluation metrics as the original publications in each dataset. For \textit{ETT} datasets, we follow Informer and use the Mean Absolute Errors (MAE)~\citep{Hyndman2006AnotherLA} and Mean Squared Errors (MSE)~\citep{Makridakis1982TheAO}. For \emph{PeMS}, we use the MAE, Root Mean Squared Errors (RMSE)  and Mean Absolute Percentage Errors (MAPE) following~\citep{Hyndman2006AnotherLA} . For long-sequence TSF tasks on \textit{Exchange Rate, Electricity and Traffic}, we follow Autoformer~\cite{xu2021autoformer} and uses MSE and MAE. For short-term point prediction on \textit{Solar-Energy, Exchange Rate, Electricity and Traffic}, following~\citep{Lai2018ModelingLA}, we use Root Relative Squared Error (RSE) and Empirical Correlation Coefficient (CORR) to evaluate the performance. The evaluation is averaged on all the predictions for multi-step predictions.

% For  MAE, MSE, RMSE, MAPE, and RSE, lower values are better. For CORR, higher values are better. 
% Moreover, in Table.~\ref{tab:datasets}, The evaluation of the multi-step task is the average error of the future steps, and the single-step task only evaluates the error of the value in the last timestamp. % $\hat{\mathbf{x}}_{t+\tau}^{K+1}$$\hat{\mathbf{X}}_{t+1:t+\tau}^{K+1}$ 

\subsection{Results and Analyses}
\label{sec:comparison_sota}
Table~\ref{tab:shortterm},~\ref{tab:longterm},~\ref{tab:etth},~\ref{tab:etth_u},~\ref{tab:traffic} provide the main experimental results of SCINet. We observe that SCINet shows superior performance than other TSF models on various tasks, including short-term, long-term and spatial-temporal time series forecasting.

\vspace{5pt}
\textbf{Short-term Time Series Forecasting:} 
we evaluate the performance of the SCINet in short-term TSF tasks with other baseline methods on \emph{Traffic}, \emph{Solar-Energy}, \emph{Electricity} and \emph{Exchange-Rate} datasets. The experimental setting is the same as~\citep{Lai2018ModelingLA}, which uses the input length of 168 to forecast different future horizons$ \left \{3, 6, 12, 24  \right \}$. 

As can be seen in Table~\ref{tab:shortterm}, the proposed SCINet outperforms existing RNN/TCN-based~ (LSTNet~\citep{Lai2018ModelingLA}, TPA-LSTM~\citep{Shih2019TemporalPA}, TCN~\citep{Bai2018AnEE}, TCN$^\dagger$) and Transformer-based~\citep{Wu2021AutoformerDT, Zhou2020InformerBE, vaswani2017attention} TSF solutions in most cases, especially for the Solar-Energy and Exchange-Rate datasets. 
%At the same time, SCINet is slightly inferior to the spatial-temporal models MTGNN~\citep{Wu2020ConnectingTD} and TPA-LSTM~\citep{Shih2019TemporalPA} in some cases. We attribute it to the fact that these datasets have relatively strong spatial relations and hence the dedicated spatial modeling architectures in these techniques contribute to their high performance. 
%
Note that, TCN$^\dagger$ denotes a variant of TCN wherein causal convolutions are replaced by normal convolutions, and improves the original TCN across all the datasets, which supports our claim in Sec.~\ref{sec:rethinking_tcn}. 
Moreover, we can also observe that the Transformer-based methods have poor performance in this task. 
%As we have mentioned in the Sec.\ref{sec:rethinking_transformer}, the Transformer-based methods aim at extracting the semantic correlations between paired elements, which can hardly learn enough temporal information from the look-back window. While 
For short-term forecasting, the recent data points 
%hold the continuous temporal correlations to the target horizon, which is 
are typically more important for accurate forecasting. However, the permutation-invariant self-attention mechanisms used in Transformer-based methods do not pay much attention to such critical information. In contrast, the general sequential models (RNN/TCN) can formulate it easily, showing quite competitive results in short-term forecasting. 
%Thus, in the experiments, even though we try to tune the hyper-parameters of the Informer, Autoformer, and vanilla Transformer, the result reported in Table~\ref{tab:shortterm} still fall far behind our SCINet. 
% Thus, we argue that the transformer architecture is merely extracting the pattern correlation rather than the actual temporal features.
% However, with limited look-back window size, the model can hardly extract enough temporal information to generate the forecasting series. This makes them more rely on the temporal information learnt from the training set. While for short-term forecasting, several transformer-based method show their drawbacks. Even though we tried to tuning the hyper-parameters of the Informer, Autoformer and vanilla Transformer, the result reported in Table~\ref{tab:shortterm} still fall far behind our SCINet. Thus, we argue that the transformer architecture is merely extracting the pattern correlation rather than the actual temporal features. 
\\ %补充原因??? 学的不是temporal feature 为什么 long-term好 short-term差 
\begin{table*}[h]
\vspace{-15pt}
\caption{Short-term forecasting performance comparison on the four datasets. The best results are shown in \textbf{bold} and second best results are highlighted with {\color[RGB]{0, 100, 148} \underline{underlined blue font}}. {\color[RGB]{230, 57, 70} IMP} shows the improvement of SCINet over the best model.}
\scriptsize
  \resizebox{\textwidth}{!}
{
\begin{tabular}{ccccllllllccccccccl}
\hline
\multicolumn{2}{c}{Model}                                                                           & \multicolumn{2}{c}{\textbf{SCINet}}                                             & \multicolumn{2}{c}{Autoformer~\cite{xu2021autoformer}}                     & \multicolumn{2}{c}{Informer~\cite{Zhou2020InformerBE}}                       & \multicolumn{2}{c}{Transformer~\cite{vaswani2017attention}}                      & \multicolumn{2}{c}{*TCN~\cite{Bai2018AnEE}}                                      & \multicolumn{2}{c}{{\color[HTML]{333333} *TCN$^{\dagger}$}}   & \multicolumn{2}{c}{LSTNet~\cite{Lai2018ModelingLA}} & \multicolumn{2}{c}{TPA-LSTM~\cite{Shih2019TemporalPA}}                                              & \multicolumn{1}{c}{\textbf{IMP}}                        \\ \hline
Metric&$\tau$                                                                      & RSE                                    & CORR                                   & \multicolumn{1}{c}{RSE} & \multicolumn{1}{c}{CORR} & \multicolumn{1}{c}{RSE} & \multicolumn{1}{c}{CORR} & \multicolumn{1}{c}{RSE} & \multicolumn{1}{c}{CORR}   & RSE                           & CORR                          & RSE                           & CORR                          & RSE          & CORR        & RSE                                 & CORR  & RSE \\ \hline
\multicolumn{1}{c|}{}                                                                          & 3  & \textbf{0.1775}                        & \textbf{0.9853}                        & \multicolumn{1}{c}{N/A} & \multicolumn{1}{c}{N/A}  & \multicolumn{1}{c}{N/A} & \multicolumn{1}{c}{N/A}  & \multicolumn{1}{c}{N/A} & \multicolumn{1}{c|}{N/A}   & {\color[HTML]{000000} 0.1940} & {\color[HTML]{000000} 0.9835} & 0.1900                        & 0.9848                        & 0.1843       & 0.9843      & {\color[RGB]{0, 100, 148} \underline{ 0.1803}} & {\color[RGB]{0, 100, 148} \underline{ 0.9850}} & {\color[RGB]{230, 57, 70} 1.55\%}    \\
\multicolumn{1}{c|}{}                                                                          & 6  & \textbf{0.2301}                        & \textbf{0.9739}                        & \multicolumn{1}{c}{N/A} & \multicolumn{1}{c}{N/A}  & \multicolumn{1}{c}{N/A} & \multicolumn{1}{c}{N/A}  & \multicolumn{1}{c}{N/A} & \multicolumn{1}{c|}{N/A}   & 0.2581                        & 0.9602                        & {\color[HTML]{000000} 0.2382} & {\color[HTML]{000000} 0.9612} & 0.2559       & 0.9690      & {\color[RGB]{0, 100, 148} \underline{ 0.2347}} & {\color[RGB]{0, 100, 148} \underline{ 0.9742}} & {\color[RGB]{230, 57, 70} 1.96\%}    \\
\multicolumn{1}{c|}{}                                                                          & 12 & \textbf{0.2997}                        & \textbf{0.9550}                        & \multicolumn{1}{c}{N/A} & \multicolumn{1}{c}{N/A}  & \multicolumn{1}{c}{N/A} & \multicolumn{1}{c}{N/A}  & \multicolumn{1}{c}{N/A} & \multicolumn{1}{c|}{N/A}   & {\color[HTML]{000000} 0.3512} & {\color[HTML]{000000} 0.9321} & 0.3353                        & 0.9432                        & 0.3254       & 0.9467      & {\color[RGB]{0, 100, 148} \underline{ 0.3234}} & {\color[RGB]{0, 100, 148} \underline{ 0.9487}} & {\color[RGB]{230, 57, 70} 7.33\%}    \\
\multicolumn{1}{c|}{\multirow{-4}{*}{Solar-Energy}}                                            & 24 & \textbf{0.4081}                        & \textbf{0.9112}                        & \multicolumn{1}{c}{N/A} & \multicolumn{1}{c}{N/A}  & \multicolumn{1}{c}{N/A} & \multicolumn{1}{c}{N/A}  & \multicolumn{1}{c}{N/A} & \multicolumn{1}{c|}{N/A}   & {\color[HTML]{000000} 0.4732} & {\color[HTML]{000000} 0.8812} & 0.4676                        & 0.8851                        & 0.4643       & 0.8870      & {\color[RGB]{0, 100, 148} \underline{ 0.4389}} & {\color[RGB]{0, 100, 148} \underline{ 0.9081}} & {\color[RGB]{230, 57, 70} 7.02\%}    \\ \hline
\multicolumn{1}{c|}{}                                                                          & 3  & {\color[HTML]{000000} \textbf{0.4216}} & {\color[HTML]{000000} \textbf{0.8920}} &0.5368 & 0.8268 & 0.5175 & 0.8515 & 0.5122  &\multicolumn{1}{l|}{0.8555} & 0.5459                        & 0.8486                        & 0.5361                        & 0.8540                        & 0.4777       & 0.8721      & {\color[RGB]{0, 100, 148} \underline{ 0.4487}} & {\color[RGB]{0, 100, 148} \underline{ 0.8812}} & {\color[RGB]{230, 57, 70} 6.04\%}    \\
\multicolumn{1}{c|}{}                                                                          & 6  & \textbf{0.4414}                        & \textbf{0.8809}                        &0.5462 & 0.8191 & 0.5258 & 0.8465 & 0.5455  &\multicolumn{1}{l|}{0.8388} & 0.6061                        & 0.8205                        & {\color[HTML]{000000} 0.5992} & {\color[HTML]{000000} 0.8197} & 0.4893       & 0.8690      & {\color[RGB]{0, 100, 148} \underline{ 0.4658}} & {\color[RGB]{0, 100, 148} \underline{ 0.8717}} & {\color[RGB]{230, 57, 70} 5.24\%}    \\
\multicolumn{1}{c|}{}                                                                          & 12 & {\color[HTML]{000000} \textbf{0.4495}} & {\color[HTML]{000000} \textbf{0.8772}} &0.5623 & 0.8082 & 0.5533 & 0.8279 & 0.5485  &\multicolumn{1}{l|}{0.8317} & 0.6367                        & 0.8048                        & 0.6061                        & 0.8205                        & 0.4950       & 0.8614      & {\color[RGB]{0, 100, 148} \underline{ 0.4641}} & {\color[RGB]{0, 100, 148} \underline{ 0.8717}} & {\color[RGB]{230, 57, 70} 3.15\%}    \\
\multicolumn{1}{c|}{\multirow{-4}{*}{Traffic}}                                                 & 24 & \textbf{0.4453}                        & \textbf{0.8825}                        &0.6020 & 0.7757 & 0.5883 & 0.8033 & 0.5934  &\multicolumn{1}{l|}{0.8048} & {\color[HTML]{000000} 0.6586} & {\color[HTML]{000000} 0.7921} & 0.6456                        & 0.7982                        & 0.4973       & 0.8588      & {\color[RGB]{0, 100, 148} \underline{ 0.4765}} & {\color[RGB]{0, 100, 148} \underline{ 0.8629}} & {\color[RGB]{230, 57, 70} 6.55\%}    \\ \hline
\multicolumn{1}{c|}{}                                                                          & 3  & \textbf{0.0740}                        & \textbf{0.9494}                        &0.1458 & 0.9032 & 0.1524 & 0.8858 & 0.1182  &\multicolumn{1}{l|}{0.9055} & {\color[HTML]{000000} 0.0892} & {\color[HTML]{000000} 0.9232} & 0.0852                        & 0.9293                        & 0.0864       & 0.9283      & {\color[RGB]{0, 100, 148} \underline{ 0.0823}} & {\color[RGB]{0, 100, 148} \underline{ 0.9439}} & {\color[RGB]{230, 57, 70} 10.09\%}   \\
\multicolumn{1}{c|}{}                                                                          & 6  & \textbf{0.0845}                        & \textbf{0.9387}                        & 0.1555 & 0.8957 & 0.1932 & 0.8660 & 0.1328  &\multicolumn{1}{l|}{0.8962} & {\color[HTML]{000000} 0.0974} & {\color[HTML]{000000} 0.9121} & 0.0924                        & 0.9235                        & 0.0931       & 0.9135      & {\color[RGB]{0, 100, 148} \underline{ 0.0916}} & {\color[RGB]{0, 100, 148} \underline{ 0.9337}} & {\color[RGB]{230, 57, 70} 7.75\%}    \\
\multicolumn{1}{c|}{}                                                                          & 12 & {\color[HTML]{000000} \textbf{0.0929}} & \textbf{0.9305}                        & 0.1541 & 0.8907 & 0.1748 & 0.8585 & 0.1375  &\multicolumn{1}{l|}{0.8849} & {\color[HTML]{000000} 0.1053} & {\color[HTML]{000000} 0.9017} & 0.0993                        & 0.9173                        & 0.1007       & 0.9077      & {\color[RGB]{0, 100, 148} \underline{ 0.0964}} & {\color[RGB]{0, 100, 148} \underline{ 0.9250}} & {\color[RGB]{230, 57, 70} 3.63\%}    \\
\multicolumn{1}{c|}{\multirow{-4}{*}{Electricity}}                                             & 24 & {\color[HTML]{000000} \textbf{0.0967}} & {\color[HTML]{000000} \textbf{0.9270}} & 0.1754 & 0.8732 & 0.2110 & 0.8347 & 0.1461  &\multicolumn{1}{l|}{0.8774} & 0.1091                        & 0.9101                        & 0.0989                        & 0.9101                        & 0.1007       & 0.9119      & {\color[RGB]{0, 100, 148} \underline{ 0.1006}} & {\color[RGB]{0, 100, 148} \underline{ 0.9133}} & {\color[RGB]{230, 57, 70} 3.88\%}    \\ \hline
\multicolumn{1}{c|}{}                                                                          & 3  & \textbf{0.0171}                        & \textbf{0.9787}                        &0.0400 & 0.9458 & 0.1392 & 0.9473 & 0.0689  &\multicolumn{1}{l|}{0.9759} & 0.0217                        & 0.9693                        & {\color[HTML]{000000} 0.0202} & {\color[HTML]{000000} 0.9712} & 0.0226       & 0.9735      & {\color[RGB]{0, 100, 148} \underline{ 0.0174}} & {\color[RGB]{0, 100, 148} \underline{ 0.979}}  & {\color[RGB]{230, 57, 70} 1.72\%}    \\
\multicolumn{1}{c|}{}                                                                          & 6  & \textbf{0.0240}                        & {\color[RGB]{0, 100, 148} \underline{ 0.9704}}    &0.0481 & 0.9197 & 0.1548 & 0.9207 & 0.0806  &\multicolumn{1}{l|}{0.9671} & 0.0263                        & 0.9633                        & {\color[HTML]{000000} 0.0257} & {\color[HTML]{000000} 0.9628} & 0.0280       & 0.9658      & {\color[RGB]{0, 100, 148} \underline{ 0.0241}} & \textbf{0.9709}                     & {\color[RGB]{230, 57, 70} 0.41\%}    \\
\multicolumn{1}{c|}{}                                                                          & 12 & \textbf{0.0331}                        & {\color[RGB]{0, 100, 148} \underline{ 0.9553}}    & 0.0638 & 0.9054 & 0.1793 & 0.8817 & 0.0893  &\multicolumn{1}{l|}{0.9476} & 0.0393                        & 0.9531                        & {\color[HTML]{000000} 0.0352} & {\color[HTML]{000000} 0.9501} & 0.0356       & 0.9511      & {\color[RGB]{0, 100, 148} \underline{ 0.0341}} & \textbf{0.9564}                     & {\color[RGB]{230, 57, 70} 2.93\%}    \\
\multicolumn{1}{c|}{\multirow{-4}{*}{\begin{tabular}[c]{@{}c@{}}Exchange\\ Rate\end{tabular}}} & 24 & \textbf{0.0436}                        & \textbf{0.9396}                        & 0.0651 & 0.8952 & 0.1998 & 0.7715 & 0.1127  &\multicolumn{1}{l|}{0.9213} & 0.0492                        & 0.9223                        & {\color[HTML]{000000} 0.0487} & {\color[HTML]{000000} 0.9314} & 0.0449       & 0.9354      & {\color[RGB]{0, 100, 148} \underline{ 0.0444}} & {\color[RGB]{0, 100, 148} \underline{ 0.9381}} & {\color[RGB]{230, 57, 70} 1.80\%}    \\ \hline
\end{tabular}
}

\label{tab:shortterm}
\begin{tablenotes} %添加此处
\tiny
{
\item - Autoformer, Informer and Transformer achieved by Autoformer~\cite{xu2021autoformer} requires pre-prossessed datasets for training. 
\item - N/A denotes no pre-prossessed dataset for training. 
\item - $*$ denotes re-implementation. \hspace{10pt} $\dagger$ denotes the variant with normal convolutions.
}
\end{tablenotes} %添加此处
\end{table*}

% \vspace{-10pt}
\textbf{Long-term Time Series Forecasting:} many real-world applications also require to predict long-term events. Therefore, we conduct the experiments on
 \textit{Exchange Rate, Electricity ,Traffic} and \textit{ETT}  datasets to evaluate the performance of SCINet on long-term TSF tasks. 
% To show our short-term forecasting performance, we conducted short-term point prediction on \textit{Solar-Energy, Exchange Rate, Electricity and Traffic}~\citep{Lai2018ModelingLA}, the results are shown in Table~\ref{tab:shortterm}. \\
In this experiment, we only compare SCINet with Transformer-based methods~\citep{Wu2021AutoformerDT, kitaev2019reformer,li2019enhancing,Zhou2020InformerBE, vaswani2017attention, Liu2022Pyraformer}, since they are more popular in recent long-term TSF research. 

As can be seen from Table~\ref{tab:longterm}, the SCINet achieves state-of-the-art performances in most benchmarks and prediction length settings. % (except ETTh1~(720)).
Overall, SCINet yields 39.89$\%$ average improvements on MSE among the above settings. In particular, for Exchange-Rate, compared to previous state-of-the-art results, SCINet gives average 65$\%$ improvements on MSE. We attribute it to that the proposed SCINet can better capture both short~(\emph{local temporal dynamics})- and long~(\emph{trend, seasonality})-term temporal dependencies to make an accurate prediction in long-term TSF. Besides, compared with the vanilla Transformer-based methods~\citep{kitaev2019reformer,li2019enhancing,Zhou2020InformerBE}, the newly-proposed Transformer-based forecasting model Autoformer~\citep{Wu2021AutoformerDT} achieves the second best performance in all experimental settings and also surpasses the SCINet in Traffic(96). This is because, Autoformer incorporates more prior knowledge about the time series data. It focuses on modelling seasonal patterns and conducts self-attention at the sub-series level~(instead of the raw data), which is much better in extracting long-term temporal patterns than vanilla Transformer-based methods.

\begin{table*}[h]
% \vspace{-10pt}
\caption{Long-term forecasting performance comparison with Transformer-based models. }
\centering
\scriptsize
\resizebox{\textwidth}{!}
{
\begin{tabular}{ccllllllllllllll|l}
\hline
\multicolumn{2}{c}{Model}                                                     & \multicolumn{2}{c}{\textbf{SCINet}}                                    & \multicolumn{2}{c}{Autoformer~\cite{Wu2021AutoformerDT}}   & \multicolumn{2}{c}{$*$Pyraformer~\cite{Liu2022Pyraformer}}                                          & \multicolumn{2}{c}{Informer~\cite{Zhou2020InformerBE}}                      & \multicolumn{2}{c}{Transformer~\cite{vaswani2017attention}}                   & \multicolumn{2}{c}{LogTrans~\cite{li2019enhancing}}                      & \multicolumn{2}{c|}{Reformer~\cite{kitaev2019reformer}} & \textbf{IMP}                               \\ \hline
\multicolumn{2}{c}{Metric}                                                                           & \multicolumn{1}{c}{MSE}            & \multicolumn{1}{c}{MAE}           & \multicolumn{1}{c}{MSE}            & \multicolumn{1}{c}{MAE} & \multicolumn{1}{c}{MSE}            & \multicolumn{1}{c}{MAE}            & \multicolumn{1}{c}{MSE} & \multicolumn{1}{c}{MAE} & \multicolumn{1}{c}{MSE} & \multicolumn{1}{c}{MAE} & \multicolumn{1}{c}{MSE} & \multicolumn{1}{c}{MAE} & \multicolumn{1}{c}{MSE}  & \multicolumn{1}{c|}{MAE}  & MSE \\ \hline
\multicolumn{1}{c|}{}                                                                          & 96  & \textbf{0.061}                     & \textbf{0.188}                    & {\color[RGB]{0, 100, 148} \underline{0.197}} & {\color[RGB]{0, 100, 148} \underline{0.323}}  & 1.748 &  1.105& 0.847                   & 0.752                   & 0.559                   & 0.587                   & 0.968                   & 0.812                   & 1.065                    & 0.829                     & {\color[RGB]{230, 57, 70} 68.98\%}   \\
\multicolumn{1}{c|}{}                                                                          & 192 & \textbf{0.106}                     & \textbf{0.244}                    & {\color[RGB]{0, 100, 148} \underline{0.300}} & {\color[RGB]{0, 100, 148} \underline{0.369}}  &  1.874 &  1.151& 1.204                   & 0.895                   & 1.168                   & 0.835                   & 1.040                   & 0.851                   & 1.188                    & 0.906                     & {\color[RGB]{230, 57, 70} 64.70\%}   \\
\multicolumn{1}{c|}{}                                                                          & 336 & \textbf{0.181}                     & \textbf{0.323}                    & {\color[RGB]{0, 100, 148} \underline{0.509}} & {\color[RGB]{0, 100, 148} \underline{0.524}} &  1.943 &  1.172 & 1.672                   & 1.036                   & 1.423                   & 0.949                   & 1.659                   & 1.081                   & 1.357                    & 0.976                     & {\color[RGB]{230, 57, 70} 64.36\%}   \\
\multicolumn{1}{c|}{\multirow{-4}{*}{\begin{tabular}[c]{@{}c@{}}Exchange\\ Rate\end{tabular}}} & 720 & \textbf{0.525}                     & \textbf{0.571}                    & {\color[RGB]{0, 100, 148} \underline{1.447}} & {\color[RGB]{0, 100, 148} \underline{0.941}}  &  2.085 &  1.206& 2.478                   & 2.478                   & 2.160                   & 1.150                   & 1.941                   & 1.127                   & 1.510                    & 1.016                     & {\color[RGB]{230, 57, 70} 63.72\%}   \\ \hline
\multicolumn{1}{c|}{}                                                                          & 96  & \textbf{0.168}                     & \textbf{0.253}                    & {\color[RGB]{0, 100, 148} \underline{0.201}} & {\color[RGB]{0, 100, 148} \underline{0.317}}  & 0.386 & 0.449 & 0.274                   & 0.368                   & 0.263                   & 0.359                   & 0.258                   & 0.357                   & 0.312                    & 0.402                     & {\color[RGB]{230, 57, 70} 16.42\%}   \\
\multicolumn{1}{c|}{}                                                                          & 192 & \textbf{0.175}                     & \textbf{0.262}                    & {\color[RGB]{0, 100, 148} \underline{0.222}} & {\color[RGB]{0, 100, 148} \underline{0.334}} &  0.378 &  0.443 & 0.296                   & 0.296                   & 0.273                   & 0.374                   & 0.266                   & 0.368                   & 0.348                    & 0.433                     & {\color[RGB]{230, 57, 70} 21.17\%}   \\
\multicolumn{1}{c|}{}                                                                          & 336 & \textbf{0.189}                     & \textbf{0.278}                    & {\color[RGB]{0, 100, 148} \underline{0.231}} & {\color[RGB]{0, 100, 148} \underline{0.338}}  &  0.376 &  0.443 & 0.300                   & 0.394                   & 0.277                   & 0.373                   & 0.280                   & 0.380                   & 0.350                    & 0.433                     & {\color[RGB]{230, 57, 70} 18.19\%}   \\
\multicolumn{1}{c|}{\multirow{-4}{*}{Electricity}}                                             & 720 & \textbf{0.231}                     & \textbf{0.316}                    & {\color[RGB]{0, 100, 148} \underline{0.254}} & {\color[RGB]{0, 100, 148} \underline{0.361}}  &  0.376 & 0.445 & 0.373                   & 0.439                   & 0.290                   & 0.378                   & 0.283                   & 0.376                   & 0.340                    & 0.420                     & {\color[RGB]{230, 57, 70} 9.06\%}    \\ \hline
\multicolumn{1}{c|}{}                                                                          & 96  &        \textbf{0.613}                            &   {\color[RGB]{0, 100, 148} \underline{0.395}}                                & \textbf{0.613} & \textbf{0.388}  & 0.867 &  0.468 & 0.719                   & 0.391                   & 0.638                   & 0.354                   & 0.684                   & 0.384                   & 0.732                    & 0.423                     & {\color[RGB]{230, 57, 70} 0.00\%}          \\
\multicolumn{1}{c|}{}                                                                          & 192 &          \textbf{0.535}                         &      \textbf{0.355}                             & {\color[RGB]{0, 100, 148} \underline{0.616}} & {\color[RGB]{0, 100, 148} \underline{0.382}}  &  0.869&  0.467 & 0.696                   & 0.379                   & 0.647                   & 0.354                   & 0.685                   & 0.390                   & 0.733                    & 0.420                     & {\color[RGB]{230, 57, 70} 13.15\%}          \\
\multicolumn{1}{c|}{}                                                                          & 336 &                   \textbf{0.540}                 &        \textbf{0.359}                           & {\color[RGB]{0, 100, 148} \underline{0.622}} & {\color[RGB]{0, 100, 148} \underline{0.337}}  &  0.881& 0.469& 0.777                   & 0.420                   & 0.669                   & 0.364                   & 0.733                   & 0.408                   & 0.742                    & 0.420                     & {\color[RGB]{230, 57, 70}13.18\% }          \\
\multicolumn{1}{c|}{\multirow{-4}{*}{Traffic}}                                                 & 720 &\textbf{0.620 }                       &        \textbf{0.394 }                        & {\color[RGB]{0, 100, 148} \underline{0.660}} & {\color[RGB]{0, 100, 148} \underline{0.408}}  &  0.896 & 0.473 & 0.864                   & 0.472                    & 0.707                   & 0.386                   & 0.717                   & 0.396                   & 0.755                    & 0.423                     & {\color[RGB]{230, 57, 70}6.06\%}          \\ \hline
\end{tabular}
}
\begin{tablenotes} %添加此处
\tiny
		\item - $*$ denotes re-implementation.
 \end{tablenotes} %添加此处
\label{tab:longterm}
\end{table*}

\begin{table*}[htbp]	
% \small
\caption{Multivariate time-series forecasting results on the \emph{ETT} datasets.}
\begin{threeparttable}
  \resizebox{\textwidth}{!}
{
\begin{tabular}{c|c|c|c|c|c|c|c|c|c|c|c|c|c|c|c|c}
\hline
                                   &                                    & \multicolumn{5}{c|}{\textbf{ETTh1}}                                                                                                                      & \multicolumn{5}{c|}{\textbf{ETTh2}}                                                                                                                      & \multicolumn{5}{c}{\textbf{ETTm1}}                                                                                                                      \\ \cline{3-17} 
                                   &                                    & \multicolumn{5}{c|}{Horizon}                                                                                                                             & \multicolumn{5}{c|}{Horizon}                                                                                                                             & \multicolumn{5}{c}{Horizon}                                                                                                                             \\ \cline{3-17} 
\multirow{-3}{*}{\textbf{Methods}} & \multirow{-3}{*}{\textbf{Metrics}} & 24                           & 48                           & 168                          & 336                          & 720                          & 24                           & 48                           & 168                          & 336                          & 720                          & 24                           & 48                           & 96                           & 288                          & 672                          \\ \hline
                                   & MSE                                & 0.686                        & 0.766                        & 1.002                        & 1.362                        & 1.397                        & 0.828                        & 1.806                        & 4.070                        & 3.875                        & 3.913                        & 0.419                        & 0.507                        & 0.768                        & 1.462                        & 1.669                        \\ %\cline{2-17} 
\multirow{-2}{*}{LogTrans~\cite{li2019enhancing}}         & MAE                                & 0.604                        & 0.757                        & 0.846                        & 0.952                        & 1.291                        & 0.750                        & 1.034                        & 1.681                        & 1.763                        & 1.552                        & 0.412                        & 0.583                        & 0.792                        & 1.320                        & 1.461                        \\ \hline
                                   & MSE                                & 0.991                        & 1.313                        & 1.824                        & 2.117                        & 2.415                        & 1.531                        & 1.871                        & 4.660                        & 4.028                        & 5.381                        & 0.724                        & 1.098                        & 1.433                        & 1.820                        & 2.187                        \\ %\cline{2-17} 
\multirow{-2}{*}{Reformer~\cite{kitaev2019reformer}}         & MAE                                & 0.754                        & 0.906                        & 1.138                        & 1.280                        & 1.520                        & 1.613                        & 1.735                        & 1.846                        & 1.688                        & 2.015                        & 0.607                        & 0.777                        & 0.945                        & 1.094                        & 1.232                        \\ \hline
                                   & MSE                                & 0.650                        & 0.702                        & 1.212                        & 1.424                        & 1.960                        & 1.143                        & 1.671                        & 4.117                        & 3.434                        & 3.963                        & 0.621                        & 1.392                        & 1.339                        & 1.740                        & 2.736                        \\ %\cline{2-17} 
\multirow{-2}{*}{LSTMa~\cite{bahdanau2015neural}}            & MAE                                & 0.624                        & 0.675                        & 0.867                        & 0.994                        & 1.322                        & 0.813                        & 1.221                        & 1.674                        & 1.549                        & 1.788                        & 0.629                        & 0.939                        & 0.913                        & 1.124                        & 1.555                        \\ \hline
                                   & MSE                                & 1.293                        & 1.456                        & 1.997                        & 2.655                        & 2.143                        & 2.742                        & 3.567                        & 3.242                        & 2.544                        & 4.625                        & 1.968                        & 1.999                        & 2.762                        & 1.257                        & 1.917                        \\ %\cline{2-17} 
\multirow{-2}{*}{LSTNet~\cite{Lai2018ModelingLA}}           & MAE                                & 0.901                        & 0.960                        & 1.214                        & 1.369                        & 1.380                        & 1.457                        & 1.687                        & 2.513                        & 2.591                        & 3.709                        & L1700                        & 1.215                        & 1.542                        & 2.076                        & 2.941                        \\ \hline
                                   & MSE                                & 0.577                        & 0.685                        & 0.931                        & 1.128                        & 1.215                        & 0.720                        & 1.457                        & 3.489                        & 2.723                        & 3.467                       & 0.323                        & 0.494                        & 0.678                        & 1.056                        & 1.192                       \\ %\cline{2-17} 
\multirow{-2}{*}{Informer~\cite{Zhou2020InformerBE}}         & MAE                                & 0.549                        & 0.625                        & 0.752                        & 0.873                        & 0.896                        & 0.665                        & 1.001                        & 1.515                        & 1.340                        &1.473                     & 0.369                        & 0.503                        & 0.614                        & 0.786                        & 0.926                     \\ \hline
                                   & MSE                                & 0.511                                             & 0.515                                             & 0.694                                             & 0.814                                             & 0.944                                             & 0.444                                             & 0.617                                             & 2.405                                             & 2.486                                             & 2.608                                             & 0.229                                             & 0.239                                             & 0.260                                              & 0.768                                             & 2.732                         \\ %\cline{2-17} 
\multirow{-2}{*}{*TCN~\cite{Bai2018AnEE}}              & MAE                               & 0.549                                             & 0.529                                             & 0.617                                             & 0.682                                             & 0.778                                             & 0.478                                             & 0.615                                             & 1.266                                             & 1.312                                             & 1.276                                             & 0.282                                             & 0.360                                              & 0.363                                             & 0.646                                             & 1.371                        \\ \hline
 & MSE & 0.479 & 0.518 & 0.758 & 0.891 & 0.963 & 0.477 & 0.934 & 3.913 & 0.907 & 0.963 & 0.332 & 0.492 & 0.543 & 0.656 & 0.901 \\
\multirow{-2}{*}{*Pyraformer~\cite{Liu2022Pyraformer}} & MAE & 0.499 & 0.520 & 0.665 & 0.738 & 0.782 & 0.537 & 0.764 & 1.557 & 0.747 & 0.783 & 0.383 & 0.475 & 0.510 & 0.598 & 0.720 \\ \hline
 & MSE & {\color[RGB]{0, 100, 148} \underline{0.406}} & {\color[RGB]{0, 100, 148} \underline{0.478}} & {\color[RGB]{0, 100, 148} \underline{0.493}} & {\color[RGB]{0, 100, 148} \underline{0.515}} & \textbf{0.499} & {\color[RGB]{0, 100, 148} \underline{0.260}} & {\color[RGB]{0, 100, 148} \underline{0.311}} & {\color[RGB]{0, 100, 148} \underline{0.466}} & {\color[RGB]{0, 100, 148} \underline{0.472}} & {\color[RGB]{0, 100, 148} \underline{0.480}} & {\color[RGB]{0, 100, 148} \underline{0.408}} & {\color[RGB]{0, 100, 148} \underline{0.499}} & {\color[RGB]{0, 100, 148} \underline{0.540}} & {\color[RGB]{0, 100, 148} \underline{0.636}} & {\color[RGB]{0, 100, 148} \underline{0.699}} \\
\multirow{-2}{*}{Autoformer~\cite{Wu2021AutoformerDT}} & MAE & {\color[RGB]{0, 100, 148} \underline{0.440}} & {\color[RGB]{0, 100, 148} \underline{0.462}} & {\color[RGB]{0, 100, 148} \underline{0.481}} & {\color[RGB]{0, 100, 148} \underline{0.492}} & \textbf{0.500} & {\color[RGB]{0, 100, 148} \underline{0.339}} & {\color[RGB]{0, 100, 148} \underline{0.372}} & {\color[RGB]{0, 100, 148} \underline{0.458}} & {\color[RGB]{0, 100, 148} \underline{0.478}} & \textbf{0.488} & {\color[RGB]{0, 100, 148} \underline{0.424}} & {\color[RGB]{0, 100, 148} \underline{0.464}} & {\color[RGB]{0, 100, 148} \underline{0.489}} & {\color[RGB]{0, 100, 148} \underline{0.533}} & {\color[RGB]{0, 100, 148} \underline{0.564}} \\ \hline
 & MSE & \textbf{0.300} & \textbf{0.361} & \textbf{0.408} & \textbf{0.504} & {\color[RGB]{0, 100, 148} \underline{0.544}} & \textbf{0.180} & \textbf{0.230} & \textbf{0.342} & \textbf{0.365} & \textbf{0.475} & \textbf{0.106} & \textbf{0.136} & \textbf{0.165} & \textbf{0.253} & \textbf{0.346} \\
\multirow{-2}{*}{SCINet} & MAE & \textbf{0.342} & \textbf{0.388} & \textbf{0.417} & \textbf{0.495} & {\color[RGB]{0, 100, 148} \underline{0.527}} & \textbf{0.263} & \textbf{0.303} & \textbf{0.380} & \textbf{0.409} & \textbf{0.488} & \textbf{0.202} & \textbf{0.230} & \textbf{0.252} & \textbf{0.315} & \textbf{0.376} \\ \hline
\textbf{IMP} & MSE           & {\color[RGB]{230, 57, 70}  26.11\%}                                          & {\color[RGB]{230, 57, 70}  24.48\%}                                             & {\color[RGB]{230, 57, 70}  17.24\%}                                           & {\color[RGB]{230, 57, 70}  2.14\%}                                             & {\color[RGB]{230, 57, 70}  -9.02\%}                                             & {\color[RGB]{230, 57, 70}  30.77\%}                                             & {\color[RGB]{230, 57, 70}  25.81\%}                                             & {\color[RGB]{230, 57, 70}  26.61\%}                                            & {\color[RGB]{230, 57, 70}  22.67\%}                                          & {\color[RGB]{230, 57, 70}  1.04\%}                                            & {\color[RGB]{230, 57, 70}  38.71\%}                                            & {\color[RGB]{230, 57, 70}  22.83\%}                                           & {\color[RGB]{230, 57, 70}  21.40\%}                                           & {\color[RGB]{230, 57, 70}  49.59\%}                                           & {\color[RGB]{230, 57, 70}  40.18\%}\\ \hline
\end{tabular}}
\begin{tablenotes} %添加此处
\tiny
		\item - $*$ denotes re-implementation. \hspace{10pt}
 \end{tablenotes} %添加此处

\end{threeparttable} %添加此处
\label{tab:etth}
\end{table*}

Besides, \textit{ETT} datasets~\citep{Zhou2020InformerBE} are originally used to evaluate the performance of long-sequence TSF tasks, which are conducted on two experimental settings, \emph{Multivariate Time-series Forecasting} and \emph{Univariate Time-series Forecasting}. For a fair comparison, we keep all input lengths $T$ the same as Informer. The results are shown in Table~\ref{tab:etth} and Table~\ref{tab:etth_u}, respectively. 

\begin{figure*}[h]	
	\subfigure[Sequence 441, Variate 3] %第一张子图
	{
		\begin{minipage}{6cm}
			\centering          %子图居中
			\includegraphics[scale=0.3]{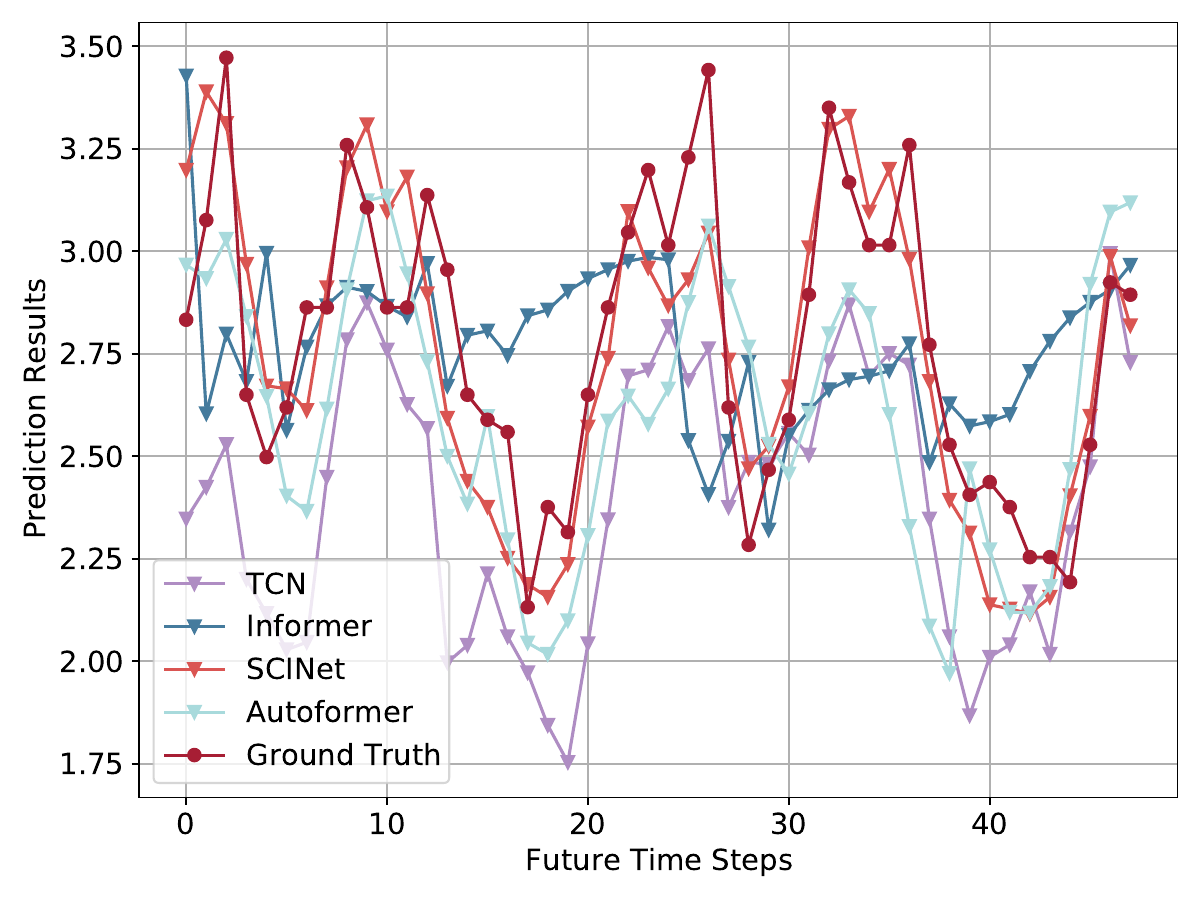}   %以pic.jpg的0.4倍大小输出
		\end{minipage}
	}
	\subfigure[Sequence 1388, Variate 1] %第二张子图
	{
		\begin{minipage}{6cm}
			\centering      %子图居中
			\includegraphics[scale=0.3]{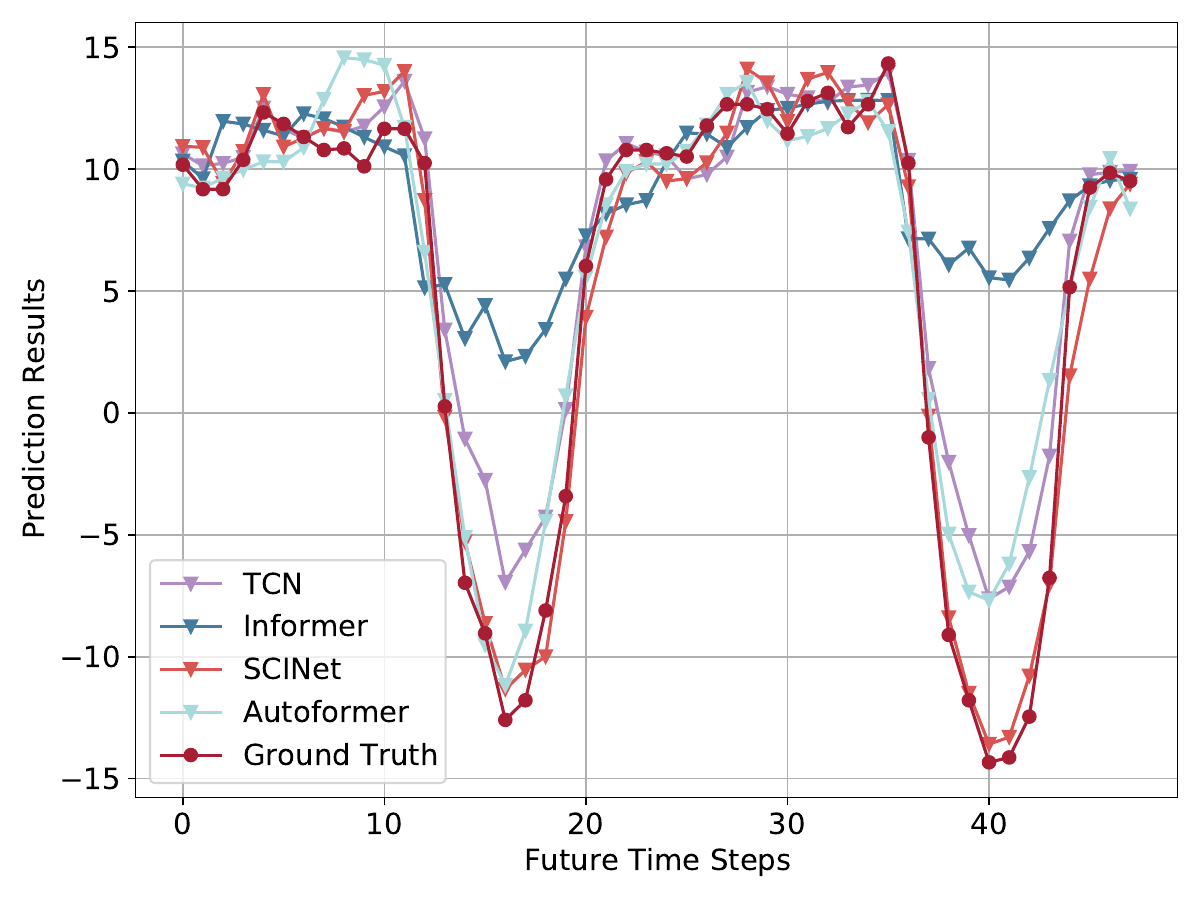}   %以pic.jpg的0.4倍大小输出
		\end{minipage}
	}
	\\
	\subfigure[Sequence 2745, Variate 4] %第二张子图
	{
		\begin{minipage}{6cm}
			\centering      %子图居中
			\includegraphics[scale=0.3]{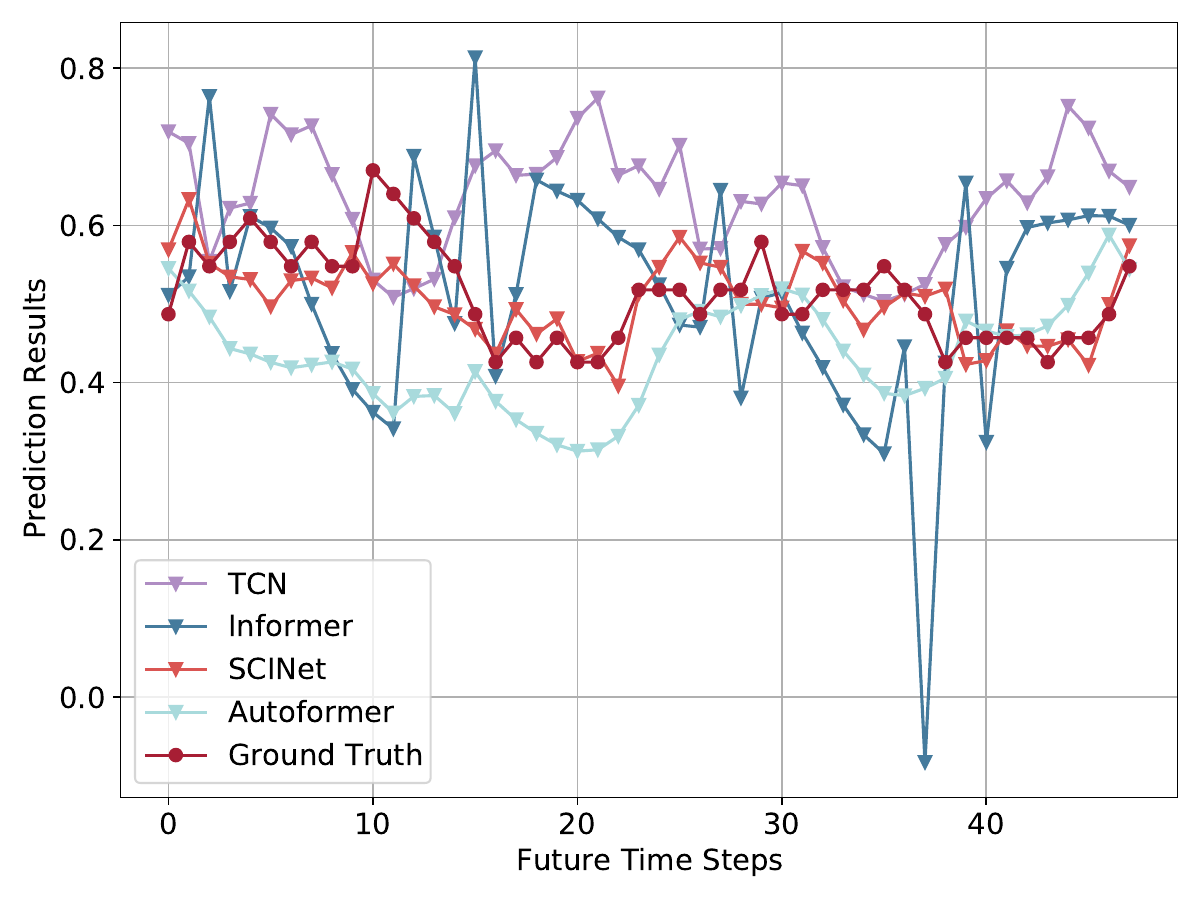}   %以pic.jpg的0.4倍大小输出
		\end{minipage}
	}
	\subfigure[Sequence 2753, Variate 5] %第二张子图
	{
		\begin{minipage}{6cm}
			\centering      %子图居中
			\includegraphics[scale=0.3]{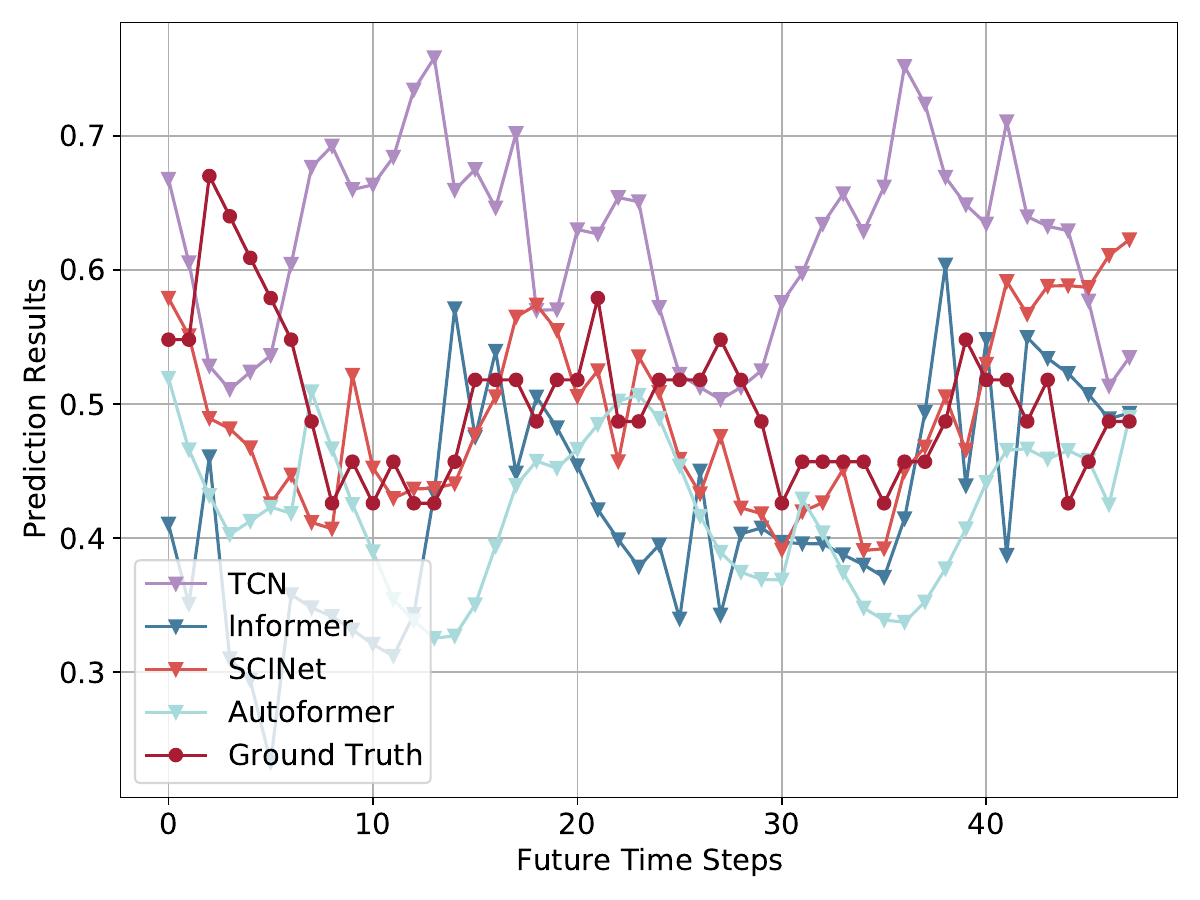}   %以pic.jpg的0.4倍大小输出
		\end{minipage}
	}

    % \vspace{-0.1cm}
	\caption{The prediction results (Horizon = $48$) of SCINet, Autoformer, Informer, and TCN on randomly-selected sequences from ETTh1 dataset. } %  %大图名称
 	\vspace{5pt}
	\label{fig:compare}  %图片引用标记
\end{figure*}

\emph{Multivariate Time-series Forecasting on ETT}: as can be seen from Table~\ref{tab:etth}, compared with RNN-based methods such as LSTMa~\citep{bahdanau2015neural} and LSTnet~\citep{Lai2018ModelingLA}, Transformer-based methods~\citep{kitaev2019reformer,li2019enhancing,Zhou2020InformerBE} are better at capturing the long-term latent patterns in the entire historical data for predicting the future, 
leading to lower prediction errors. 
However, TCN
% -based methods~\citep{Bai2018AnEE, Wu2020ConnectingTD} 
further outperforms such vanilla Transformer-based methods~\citep{kitaev2019reformer,li2019enhancing,Zhou2020InformerBE}, because the stacked convolutional layers allow for more effective local-to-global temporal relation learning for multivariate time series.
% compared with RNN and vanilla Transformer-based methods.
% Specially, TCN$^\dagger$ denotes a variant of TCN wherein causal convolutions are replaced by normal convolutions, and improves the original TCN across all the \textit{ETT} datasets. 
It is worth noting that SCINet outperforms all the above models by a large margin. Fig.~\ref{fig:compare} presents the qualitative results on some randomly selected sequences of the ETTh1 dataset, which clearly demonstrate the capability of SCINet in obtaining the trend and seasonality of time series for TSF. 

\emph{Univariate Time-series Forecasting on ETT}: 
in this experimental setting, we bring several strong baseline methods for univariate forecasting into comparison, including ARIMA, Prophet~\citep{Taylor2018ForecastingAS}, DeepAR~\citep{salinas2020deepar} and N-Beats~\citep{ Oreshkin2020NBeats}. 
In Table~\ref{tab:etth_u}, we can observe that N-Beats is superior to other baseline methods in most cases. In fact, N-Beats also takes the unique properties of time series into consideration and directly learns a trend and a seasonality model using a very deep stack of fully-connected layers with residuals, which is a departure from the predominant architectures, such as RNNs, CNNs and Transformers. Nevertheless, the performance of SCINet is still much better than N-Beats. 

We attribute the significant performance improvements of SCINet on the ETT datasets to: (i) SCINet effectively captures temporal dependencies from multiple temporal resolutions; (ii) \textit{ETT} datasets are publicly available recently and domain-specific solutions tuned specifically for these datasets do not exist yet. 

\begin{table*}[htbp]
\caption{Univariate time-series forecasting results on the \emph{ETT} datasets.}
\resizebox{\textwidth}{!}
{
\begin{tabular}{c|c|c|c|c|c|c|c|c|c|c|c|c|c|c|c|c}
\hline
\multirow{3}{*}{\textbf{Methods}} & \multirow{3}{*}{\textbf{Metrics}} & \multicolumn{5}{c|}{\textbf{ETTh1}}                                                                              & \multicolumn{5}{c|}{\textbf{ETTh2}}                                                                              & \multicolumn{5}{c}{\textbf{ETTm1}}                                                                              \\ \cline{3-17} 
                                  &                                   & \multicolumn{5}{c|}{Horizon}                                                                                     & \multicolumn{5}{c|}{Horizon}                                                                                     & \multicolumn{5}{c}{Horizon}                                                                                     \\ \cline{3-17} 
                                  &                                   & 24                   & 48                   & 168                  & 336                  & 720                  & 24                   & 48                   & 168                  & 336                  & 720                  & 24                   & 48                   & 96                   & 288                  & 672                  \\ \hline

\multirow{2}{*}{ARIMA}            & MSE                               & 0.108                & 0.175                & 0.396                & 0.468                & 0.659                & 3.554                & 3.190                & 2.800                & 2.753                & 2.878                & 0.090                & 0.179                & 0.272                & 0.462                & 0.639                \\  
                                  & MAE                               & 0.284                & 0.424                & 0.504                & 0.593                & 0.766                & 0.445                & 0.474                & 0.595                & 0.738                & 1.044                & 0.206                & 0.306                & 0.399                & 0.558                & 0.697                \\ \hline
\multirow{2}{*}{Prophet~\cite{Taylor2018ForecastingAS}}          & MSE                               & 0.115                & 0.168                & 1.224                & 1.549                & 2.735                & 0.199                & 0.304                & 2.145                & 2.096                & 3.355                & 0.120                & 0.133                & 0.194                & 0.452                & 2.747                \\  
                                  & MAE                               & 0.275                & 0.330                & 0.763                & 1.820                & 3.253                & 0.381                & 0.462                & 1.068                & 2.543                & 4.664                & 0.290                & 0.305                & 0.396                & 0.574                & 1.174                \\ \hline
\multirow{2}{*}{DeepAR~\cite{salinas2020deepar}}           & MSE                               & 0.107                & 0.162                & 0.239                & 0.445                & 0.658                & 0.098                & 0.163                & 0.255                & 0.604                & 0.429                & 0.091                & 0.219                & 0.364                & 0.948                & 2.437                \\  
                                  & MAE                               & 0.280                & 0.327                &0.422                & 0.552                & 0.707                & 0.263                & 0.341                & 0.414                & 0.607                & 0.580                & 0.243                & 0.362                & 0.496                & 0.795                & 1.352                \\ \hline
\multirow{2}{*}{N-Beats~\cite{ Oreshkin2020NBeats}}            & MSE                               & 0.042                & 0.065              & 0.106                & 0.127                & 0.269               & 0.078                & 0.123                & 0.244                & 0.270                & 0.281                & 0.031                & 0.056               & 0.095                & 0.157               & 0.207               \\  
                                                & MAE                               & 0.156                & 0.200                & 0.255               & 0.284                & 0.422                & 0.210                & 0.271                & 0.393              & 0.418                & 0.432                & 0.117                & 0.168               & 0.234              & 0.311                & 0.370                \\ \hline

\multirow{2}{*}{Informer~\cite{Zhou2020InformerBE}}         & MSE                               & 0.098                &0.158 &0.183 & 0.222                & 0.269                &0.093 &0.155 &0.232 & 0.263                &0.277 &0.030 & 0.069                & 0.194                &0.401 &0.512 \\  
                                  & MAE                               & 0.247                &0.319 &0.346 & 0.387                & 0.435                &0.240 &0.314 &0.389 & 0.417                &0.431 &0.137 & 0.203                & 0.372                & 0.554                &0.644 \\ \hline
& MSE & {\color[RGB]{0, 100, 148} \underline{0.057}} & {\color[RGB]{0, 100, 148} \underline{0.103}} & {\color[RGB]{0, 100, 148} \underline{0.090}} & {\color[RGB]{0, 100, 148} \underline{0.106}} & {\color[RGB]{0, 100, 148} \underline{0.120}} & {\color[RGB]{0, 100, 148} \underline{0.110}} & {\color[RGB]{0, 100, 148} \underline{0.123}} & {\color[RGB]{0, 100, 148} \underline{0.188}} & {\color[RGB]{0, 100, 148} \underline{0.225}} & \textbf{0.257} & {\color[RGB]{0, 100, 148} \underline{0.025}} & \textbf{0.039} & \textbf{0.057} & \textbf{0.103} & \textbf{0.110} \\
\multirow{-2}{*}{Autoformer~\cite{Wu2021AutoformerDT}} & MAE & {\color[RGB]{0, 100, 148} \underline{0.188}} & {\color[RGB]{0, 100, 148} \underline{0.257}} & {\color[RGB]{0, 100, 148} \underline{0.235}} & {\color[RGB]{0, 100, 148} \underline{0.254}} & {\color[RGB]{0, 100, 148} \underline{0.277}} & {\color[RGB]{0, 100, 148} \underline{0.259}} & {\color[RGB]{0, 100, 148} \underline{0.271}} & {\color[RGB]{0, 100, 148} \underline{0.340}} & {\color[RGB]{0, 100, 148} \underline{0.376}} & \textbf{0.402} & {\color[RGB]{0, 100, 148} \underline{0.122}} & \textbf{0.156} & \textbf{0.184} & \textbf{0.253} & \textbf{0.261} \\ \hline
 & MSE & \textbf{0.029} & \textbf{0.041} & \textbf{0.071} & \textbf{0.084} & \textbf{0.099} & 
 \textbf{0.065} & \textbf{0.093} & \textbf{0.158} & \textbf{0.166} & {\color[RGB]{0, 100, 148} \underline{0.286}} &
 \textbf{0.019} & {\color[RGB]{0, 100, 148} \underline{0.045}} & {\color[RGB]{0, 100, 148} \underline{0.064}} & {\color[RGB]{0, 100, 148} \underline{0.111}} & {\color[RGB]{0, 100, 148} \underline{0.165}} \\
\multirow{-2}{*}{SCINet} & MAE & \textbf{0.127} & \textbf{0.154} & \textbf{0.210} & \textbf{0.234} & \textbf{0.250} & 
\textbf{0.183} & \textbf{0.227} & \textbf{0.311} & \textbf{0.329} & {\color[RGB]{0, 100, 148} \underline{0.429}} &
 \textbf{0.084} & {\color[RGB]{0, 100, 148} \underline{0.138}} & {\color[RGB]{0, 100, 148} \underline{0.183}} & {\color[RGB]{0, 100, 148} \underline{0.252}} & {\color[RGB]{0, 100, 148} \underline{0.316}} \\ \hline
\multirow{1}{*}{\textbf{IMP}}              & MSE                               & {\color[RGB]{230, 57, 70} 49.12\%}                & {\color[RGB]{230, 57, 70} 60.19\%}                & {\color[RGB]{230, 57, 70} 21.11\%}                & {\color[RGB]{230, 57, 70} 20.75\%}                & {\color[RGB]{230, 57, 70} 17.50\%}              & {\color[RGB]{230, 57, 70} 40.90\%}                & {\color[RGB]{230, 57, 70} 24.39\%}                & {\color[RGB]{230, 57, 70} 15.96\%}               & {\color[RGB]{230, 57, 70} 26.22\%}               & {\color[RGB]{230, 57, 70} -11.28\%}               & {\color[RGB]{230, 57, 70} 24.00\%}                & {\color[RGB]{230, 57, 70} -15.38\%}                & {\color[RGB]{230, 57, 70} -12.28\%}                & {\color[RGB]{230, 57, 70} -7.76\%}               &  {\color[RGB]{230, 57, 70} -50.00\%}                    \\\hline
\end{tabular}}
%\begin{tablenotes} %添加此处
%\tiny
		%\item   $*$ Source codes of the N-Beats are downloaded from the %\url{https://github.com/philipperemy/N-Beats}. 
% \end{tablenotes} %添加此处
\label{tab:etth_u}
%\vspace{-8pt}
\end{table*}

\vspace{3pt}
\textbf{Spatial-temporal Time Series Forecasting:} besides the general TSF tasks, there is also a large category of data related to spatial-temporal forecasting. For example, traffic datasets \emph{PeMS}~\citep{Chen2001FreewayPM} (PEMS03,
PEMS04, PEMS07 and PEMS08) are complicated spatial-temporal time series for public traffic network and they have been investigated for decades. Most recent approaches: DCRNN~\cite{li2018diffusion}, STGCN~\cite{Yu2018SpatioTemporalGC}, ASTGCN~\cite{Guo2019AttentionBS}, GraphWaveNet~\cite{Wu2019GraphWF}, STSGCN~\cite{Song2020SpatialTemporalSG}, AGCRN~\cite{bai2020adaptive}, LSGCN~\cite{Huang2020LSGCNLS} and STFGNN~\cite{li2021spatial} use graph neural networks to capture spatial relations while modeling temporal dependencies via conventional TCN or LSTM architectures. We follow the same experimental settings as the above works. As shown in Table~\ref{tab:traffic}, these GNN-based methods generally perform better than pure RNN or TCN-based methods. However, SCINet still achieves better performance without sophisticated spatial relation modelling, which further proves the superb temporal modeling capabilities of SCINet.

\begin{table*}[h]	
%~\citep{li2018diffusion,Yu2018SpatioTemporalGC,Wu2019GraphWF,Song2020SpatialTemporalSG,mengzhang2020spatial,Bai2020AdaptiveGC,Huang2020LSGCNLS}
% \vspace{-10pt}
\caption{Performance comparison of different approaches on the \emph{PeMS} datasets.}
\begin{threeparttable} %添加此处
% \vspace{-5pt}
  \resizebox{\textwidth}{!}
{
\begin{tabular}{c|c|c|c|c|c|c|c|c|c|c|c|c|c|c}
\hline
                                    &                                    & \multicolumn{12}{c|}{\textbf{Methods}}                                                                                                                                                         & \textbf{IMP}                      \\ \cline{3-15} 
\multirow{-2}{*}{\textbf{Datasets}} & \multirow{-2}{*}{\textbf{Metrics}} & *LSTM  & *TCN  & *TCN$^{\dagger}$ & DCRNN & STGCN & ASTGCN(r) & GraphWaveNet & STSGCN & STFGNN                                     & AGCRN                                       & LSGCN & SCINet         & MAE          \\ \hline
                                    & MAE                                & 21.33 & 19.32 & 18.87 & 18.18 & 17.49 & 17.69     & 19.85        & 17.48  & 16.77                                       & {\color[RGB]{0, 100, 148} \underline{ *15.98}} & -     & \textbf{14.98} & {\color[RGB]{230, 57, 70}  6.26\%}  \\ %\cline{2-15} 
                                    & MAPE                               & 21.33 & 19.93 & 18.63 & 18.91 & 17.15 & 19.40     & 19.31        & 16.78  & 16.30                                       & {\color[RGB]{0, 100, 148} \underline{ *15.23}} & -     & \textbf{14.11} & {\color[RGB]{230, 57, 70}  7.36\%}  \\ %\cline{2-15} 
\multirow{-3}{*}{\textbf{PEMS03}}   & RMSE                               & 35.11 & 33.55 & 32.24 & 30.31 & 30.12 & 29.66     & 32.94        & 29.21  & 28.34 & {\color[RGB]{0, 100, 148} \underline{ *28.25 }}                                      & -     & \textbf{24.08} & {\color[RGB]{230, 57, 70}  8.37\%}  \\ \hline
                                    & MAE                                & 25.14 & 23.22 & 22.81 & 24.70 & 22.70 & 22.93     & 25.45        & 21.19  & 19.83                                    & {\color[RGB]{0, 100, 148} \underline{ 19.83}} & 21.53 & \textbf{18.95} & {\color[RGB]{230, 57, 70}  4.44\%}  \\ %\cline{2-15} 
                                    & MAPE                               & 20.33 & 15.59 & 14.31 & 17.12 & 14.59 & 16.56     & 17.29        & 13.90  & 13.02                                       & {\color[RGB]{0, 100, 148} \underline{ 12.97}} & 13.18 & \textbf{11.86} & {\color[RGB]{230, 57, 70} 8.56\%} \\ %\cline{2-15} 
\multirow{-3}{*}{\textbf{PEMS04}}   & RMSE                               & 39.59 & 37.26 & 36.87 & 38.12 & 35.55 & 35.22     & 39.70        & 33.65  & {\color[RGB]{0, 100, 148} \underline{ 31.88}}                                       & 32.30 & 33.86 & \textbf{30.89} & {\color[RGB]{230, 57, 70}  4.40\%}  \\ \hline
                                    & MAE                                & 29.98 & 32.72 & 30.53 & 28.30 & 25.38 & 28.05     & 26.85        & 24.26  & {\color[RGB]{0, 100, 148} \underline{22.07}}                                      & *22.37 & -     & \textbf{21.19} & {\color[RGB]{230, 57, 70}  5.27\%}  \\ %\cline{2-15} 
                                    & MAPE                               & 15.33 & 14.26 & 13.88 & 11.66 & 11.08 & 13.92     & 12.12        & 10.21  & 9.21                                        & {\color[RGB]{0, 100, 148} \underline{ *9.12}}  & -     & \textbf{8.83}  & {\color[RGB]{230, 57, 70}  3.18\%}  \\ %\cline{2-15} 
\multirow{-3}{*}{\textbf{PEMS07}}   & RMSE                               & 42.84 & 42.23 & 41.02 & 38.58 & 38.78 & 42.57     & 42.78        & 39.03  &{\color[RGB]{0, 100, 148} \underline{35.80}}                                       & *36.55 & -     & \textbf{34.03} & {\color[RGB]{230, 57, 70}  6.89\%}  \\ \hline
                                    & MAE                                & 22.20 & 22.72 & 21.42 & 17.86 & 18.02 & 18.61     & 19.13        & 17.13  & 16.64                                       & {\color[RGB]{0, 100, 148} \underline{15.95}} & 17.73 & \textbf{15.72} & {\color[RGB]{230, 57, 70} 1.44\%}  \\ %\cline{2-15} 
                                    & MAPE                               & 15.32 & 14.03 & 13.09 & 11.45 & 11.40 & 13.08     & 12.68        & 10.96  & 10.60                                       & {\color[RGB]{0, 100, 148} \underline{10.09}} & 11.20 & \textbf{9.80}  & {\color[RGB]{230, 57, 70}  2.87\%}  \\ %\cline{2-15} 
\multirow{-3}{*}{\textbf{PEMS08}}   & RMSE                               & 32.06 & 35.79 & 34.03 & 27.83 & 27.83 & 28.16     & 31.05        & 26.80  & 26.22                                       & {\color[RGB]{0, 100, 148} \underline{25.22}} & 26.76 & \textbf{24.76} & {\color[RGB]{230, 57, 70}  1.82\%}  \\ \hline
\end{tabular}}
\begin{tablenotes} %添加此处
\tiny
		\item  - dash denotes that the methods do not implement on this dataset. \hspace{2pt} $*$ denotes re-implementation or re-training. \hspace{2pt} $\dagger$ denotes the variant with normal convolutions. %\\
%		The results of DCRNN and ASTGCN are extracted from \cite{mengzhang2020spatial} 
 \end{tablenotes} %添加此处

\end{threeparttable} %添加此处
\label{tab:traffic}
\end{table*}

%We attribute it to the fact that the strong temporal correlations among the traffic data play a more important role than the spatial correlations for the forecasting task and SCINet excels in temporal modeling. 

% The visualization results in Fig.~\ref{fig:compare} (c) also show that SCINet captures more details than the second best model, AGCRN~\citep{Bai2020AdaptiveGC}.
%\red{\textbf{Predictability analysis of SCINet.}} 
%As discussed earlier, SCINet learns an enhanced sequence representation and feed it into the fully-connected layer to yield the final prediction. 
%To figure out what benefits the enhanced representation for prediction, 
\vspace{3pt}
\textbf{Predictability estimation:} inspired by~\citep{Huang2019EnhancedTS,Pennekamp2019TheIP}, we use \textit{permutation entropy} (\textit{PE})~\citep{Bandt2002PermutationEA} to measure the predictability of the original input and the enhanced representation learnt by SCINet. Time series with lower PE values are regarded as less complex, thus theoretically easier to predict\footnote{Please note that PE is only a quantitative measurement based on complexity. It would not be proper to say that a time series with lower PE value will be always easier to predict than a different type of time series with a higher PE value because the prediction accuracy also depends on many other factors, such as the available data for training, the trend and seasonality elements of the time series data, as well as the predictive model.}. The PE values of the original time series and the corresponding enhanced representations are shown in Table~\ref{tab:PE}.  

\begin{table*}[h]
% \vspace{-0.3cm}
\caption{Permutation entropy comparison before and after SCINet.}
\begin{threeparttable}
% \small
\resizebox{\textwidth}{!}{
\begin{tabular}{c|c|c|c|c|c|c|c|c|c|c}
\hline
\multicolumn{2}{c|}{\multirow{2}{*}{Permutation Entropy}} & \multicolumn{9}{c}{Datasets}                                                                 \\ \cline{3-11} 
\multicolumn{2}{c|}{}                                     & ETTh1   & Traffic & Solar-Energy & Electricity & Exc-rate & PEMS03 & PEMS04 & PEMS07 & PEMS08 \\ \hline
Parameters                  & m ($\tau$ = 1)$^*$          & 6       & 6       & 7            & 6           & 6        & 6      & 6      & 6      & 6      \\ \hline
\multirow{2}{*}{Value}      & Original Input              & 0.8878  & 0.9371  & 0.4739       & 0.9489      & 0.8260   & 0.9649& 0.9203       & 0.9148 & 0.9390  \\ \cline{2-11} 
                            & Enhanced Representation     &0.7096   &  0.8832       &   0.3537           &   0.8901          & 0.7836          & 0.8377       & 0.8749       &  0.8330      & 0.8831       \\ \hline
\end{tabular}}
\begin{tablenotes} %添加此处
\tiny
		\item  $^*$ $m$ (embedding dimension) and $\tau$ (time-lag) are two parameters used for calculating PE, and the values are selected following ~\citep{Pennekamp2019TheIP,Huang2019EnhancedTS}.
     \end{tablenotes} %添加此处

\end{threeparttable} %添加此处
\label{tab:PE}
% \vspace{-0.4cm}
\end{table*}

As can be observed, the enhanced representations learnt by SCINet indeed have lower PE values compared with the original inputs, which indicates that it is easier to predict the future from the enhanced representations using the same forecaster.

% \vspace{5pt}
% \textbf{Visualization:}
% we present the qualitative results on some randomly selected sequences of the ETTh1 dataset in Fig.~\ref{fig:compare}, which clearly shows the superior performance of SCINet in obtaining the trend and seasonality of time series.

% outperforms all the above models by a large margin in most scenario. Fig.~\ref{fig:compare} presents the qualitative results on some randomly selected sequences, which clearly demonstrate the capability of SCINet in obtaining the trend and seasonality of time series for TSF.
\newpage
\subsection{Ablation studies}
\label{sec:ab}

To evaluate the impact of each main component used in SCINet, we experiment on several model variants on two datasets: \emph{ETTh1} and \emph{PEMS08}. % \red{We set $L\!=\!3$, $K\!=\!1$ and the kernal size of the convolution in $\phi$, $\rho$, $\psi$, and $\eta$ to 5 for all the baselines in this section, the details of the structure are put in the Appendix. }
% The kernal size the convolution in $\phi$, $\rho$, $\psi$, and $\eta$ are set to be $5$ for all the baselines in this section.
% The details of the $\phi$, $\rho$, $\psi$, and $\eta$ are put in the Appendix.
% is put the experiments of other $L$ and $K$ values in the Appendix. 
% The results are shown in Fig.~\ref{fig:ab}.
\vspace{5pt}

\textbf{SCIBlock}: we first set the number of stacks $K\!=\!1$ and the number of SCINet levels $L\!=\!3$ . 
For the SCI-Block design, to validate the effectiveness of the interactive learning and the distinct convolution weights for processing the sub-sequences, we experiment on two variants, namely \emph{w/o. InterLearn} and \emph{WeightShare}. 
The \emph{w/o. InterLearn} is obtained by removing the interactive-learning procedure described in Eq.~(\ref{eq:inter_1}) and~(\ref{eq:inter_2}). In this case, the two sub-sequences would be updated using $\mathbf{F}^{'}_{odd}\!=\!\rho(\phi (\mathbf{F}_{odd})) $ and $\mathbf{F}^{'}_{even}\!=\! \eta(\psi (\mathbf{F}_{even})) $. 
For \emph{WeightShare}, the modules $\phi$, $\rho$, $\psi$, and $\eta$ share the same weight. 

\begin{figure}[t]	
\centering
\includegraphics[width=0.9\textwidth]{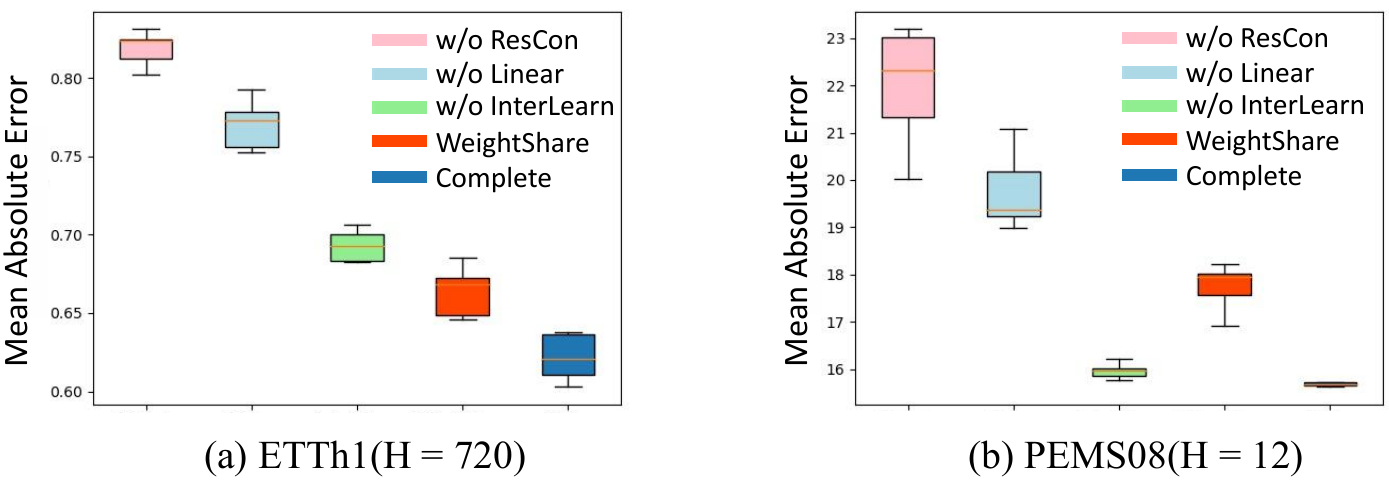}
\caption{Component analysis of SCINet on two datasets. Smaller values are better. See Section~\ref{sec:ab}.}
\label{fig:ab}
\end{figure}

The evaluation results in Fig.~\ref{fig:ab} show that both interactive learning and distinct weights are essential, as they improve the prediction accuracies of both datasets at various prediction horizons. At the same time, comparing Fig.~\ref{fig:ab}(a) with Fig.~\ref{fig:ab}(b), we can observe that interactive learning is more effective for cases with longer look-back window sizes. This is because, intuitively, we can extract more effective features by exchanging information between the downsampled sub-sequences when there are longer look-back windows for such interactions.
%heterogeneous information with the different convolutional kernels for down-sampled sub-sequences with longer look-back window sizes. However, if the information cannot be fused in time, it will result in the information block among different temporal resolutions and restrict the representation learning capability. Hence, the interaction plays a more critical role in prediction.}
     
\vspace{5pt}
\textbf{SCINet}: for the design of SCINet with multiple levels of SCI-Blocks, we also experiment on two variants.
% \textit{w.r.t.} the residual connection and the final fully-connected layer, respectively.
The first variant \emph{w/o. ResConn} is obtained by removing the residual connection from the complete SCINet. 
The other variant \emph{w/o. Linear} removes the decoder (i.e., the fully-connected layer) from the complete model. 
As can be observed in Fig.~\ref{fig:ab}, removing the residual connection leads to a significant performance drop. Besides the general benefit in facilitating the model training, more importantly, the predictability of the original time series is enhanced with the help of residuals.
% indicating the importance of the residual learning to ease the prediction. 
The fully-connected layer is also critical for prediction accuracy, indicating the effectiveness of the decoder in extracting and fusing the most relevant temporal information according to the given supervision for prediction.

We also conduct comprehensive ablation studies on the impact of $K$ (number of stacks) and $L$ (number of levels), and the selection of operator in the interact learning mechanism. These results are shown in the Appendix ~\ref{appendix:KL} due to space limitation.

% \red{Fig. (a) move the + and _ up a little; Fig. (c) replace rectangle with dashed line under concat.}

\section{Limitations and Future Work}
\label{sec:limation_fw}
In this paper, we mainly focus on TSF problem for the \textit{regular time series} collected at even intervals of time and ordered chronologically. However, in real-world applications, the time series might contain noisy data, missing data or collected at irregular time intervals, which is referred to as \textit{irregular time series}. The proposed SCINet is relatively robust to the noisy data thanks to the progressive downsampling and interactive learning procedure, but it might be affected by the missing data if the ratio exceeds a certain threshold, wherein the downsampling-based multi-resolution sequence representation in SCINet may introduce biases, leading to poor prediction performance. The proposed downsampling mechanism may also have difficulty handling data collected at irregular intervals. We plan to take the above issues into consideration in the future development of SCINet. 

Moreover, this work focuses on the deterministic time series forecasting problem. Many application scenarios require probabilistic forecasts, and we plan to revise SCINet to generate such prediction results. 

Finally, while SCINet generates promising results for spatial-temporal time series without explicitly modeling spatial relations, the forecasting accuracy could be further improved by incorporating dedicated spatial models. We plan to investigate such solutions in our future work.  

%To extend the application scenarios of SCINet, e.g., for handling irregular time series, in our future work, we plan to combine it with other data imputation techniques to better model the temporal dependencies of the missing data.

\section{Conclusion}
In this paper, we propose a novel neural network architecture, sample convolution and interaction network (\textit{SCINet}) for time series modeling and forecasting, motivated by the unique properties of time series data compared to generic sequence data. The proposed SCINet is a hierarchical downsample-convolve-interact structure with a rich set of convolutional filters. It iteratively extracts and exchanges information at different temporal resolutions and learns an effective representation with enhanced predictability. Extensive experiments on various real-world TSF datasets demonstrate the superiority of our model over state-of-the-art methods.
%The basic building block, \textit{SCI-Block}, downsamples the input data/feature into two sub-sequences and performs feature extraction and interaction to preserve the heterogeneity information of each sub-part and compensate for the information loss during the downsampling procedure. 

\begin{ack}
% \textbf{Acknowledgements: }
This work was supported in part by Alibaba Group Holding Ltd. under Grant No. TA2015393. We thank the anonymous reviewers for their constructive comments and suggestions.
\end{ack}

\def\bibfont{\fontsize{8.3}{10.3}\selectfont}
% \bibliographystyle{ACM-Reference-Format}
{\small
\bibliographystyle{ieee_fullname}
\bibliography{sample-base}
}


% \appendix

\newpage

%%%%%%%%%%%%%%%%%%%%%%%%%%%%%%%%%%%%%%%%%%%%%%%%%%%%%%%%%%%%
\section*{Checklist}

%%% BEGIN INSTRUCTIONS %%%
The checklist follows the references.  Please
read the checklist guidelines carefully for information on how to answer these
questions.  For each question, change the default \answerTODO{} to \answerYes{},
\answerNo{}, or \answerNA{}.  You are strongly encouraged to include a {\bf
justification to your answer}, either by referencing the appropriate section of
your paper or providing a brief inline description.  For example:
\begin{itemize}
  \item Did you include the license to the code and datasets? \answerYes{See Section~\ref{gen_inst}.}
  \item Did you include the license to the code and datasets? \answerNo{The code and the data are proprietary.}
  \item Did you include the license to the code and datasets? \answerNA{}
\end{itemize}
Please do not modify the questions and only use the provided macros for your
answers.  Note that the Checklist section does not count towards the page
limit.  In your paper, please delete this instructions block and only keep the
Checklist section heading above along with the questions/answers below.
%%% END INSTRUCTIONS %%%

\begin{enumerate}

\item For all authors...
\begin{enumerate}
  \item Do the main claims made in the abstract and introduction accurately reflect the paper's contributions and scope?
    \answerYes{}
  \item Did you describe the limitations of your work?
    \answerYes{} See section \ref{sec:limation_fw}
  \item Did you discuss any potential negative societal impacts of your work?
    \answerNo{}
  \item Have you read the ethics review guidelines and ensured that your paper conforms to them?
    \answerYes{}
\end{enumerate}

\item If you are including theoretical results...
\begin{enumerate}
  \item Did you state the full set of assumptions of all theoretical results?
    \answerNA{}
        \item Did you include complete proofs of all theoretical results?
    \answerNA{}
\end{enumerate}

\item If you ran experiments...
\begin{enumerate}
  \item Did you include the code, data, and instructions needed to reproduce the main experimental results (either in the supplemental material or as a URL)?
    \answerYes{} See the Appendix and link in abstract. 
  \item Did you specify all the training details (e.g., data splits, hyperparameters, how they were chosen)?
    \answerYes{} See the Appendix
        \item Did you report error bars (e.g., with respect to the random seed after running experiments multiple times)?
    \answerNo{}
        \item Did you include the total amount of compute and the type of resources used (e.g., type of GPUs, internal cluster, or cloud provider)?
    \answerYes{} See the Appendix
\end{enumerate}

\item If you are using existing assets (e.g., code, data, models) or curating/releasing new assets...
\begin{enumerate}
  \item If your work uses existing assets, did you cite the creators?
    \answerYes{} Detail seen in Appendix. 
  \item Did you mention the license of the assets?
    \answerYes{}
  \item Did you include any new assets either in the supplemental material or as a URL?
    \answerNo{}
  \item Did you discuss whether and how consent was obtained from people whose data you're using/curating?
    \answerNA{}
  \item Did you discuss whether the data you are using/curating contains personally identifiable information or offensive content?
    \answerNA{}
\end{enumerate}

\item If you used crowdsourcing or conducted research with human subjects...
\begin{enumerate}
  \item Did you include the full text of instructions given to participants and screenshots, if applicable?
    \answerNA{}
  \item Did you describe any potential participant risks, with links to Institutional Review Board (IRB) approvals, if applicable?
    \answerNA{}
  \item Did you include the estimated hourly wage paid to participants and the total amount spent on participant compensation?
    \answerNA{}
\end{enumerate}

\end{enumerate}

\appendix
\section*{Appendix}

\setcounter{secnumdepth}{2}

In this appendix, we first introduce the datasets and evaluation metrics used in the experiments in Section~\ref{sec:Evaluation}. Then,
we provide extra experimental results in Section~\ref{sec:extra_results}. In Section~\ref{appendix:reprod}, we present details of network design, training scheme, and hyper-parameter tuning. 

\section{Datasets and Evaluation Metrics}
\label{sec:Evaluation}

We conduct experiments on $11$ popular time series datasets: (1)~\emph{Electricity Transformer Temperature}~\citep{Zhou2020InformerBE}~(ETTh(1,2),ETTm1) \footnote{\url{https://github.com/zhouhaoyi/ETDataset}}consists of $2$ year electric power data collected from two separated counties of China. Each data point includes an "oil temperature" value and $6$ power load features.
(2)~\emph{Traffic}\footnote{\url{http://pems.dot.ca.gov}} contains the hourly data describing the road occupancy rates (ranging from $0$ to $1$) that are recorded by the sensors on San Francisco Bay area freeways from $2015$ to $2016$ ($48$ months in total). 
(3)~\emph{Solar-Energy}\footnote{\url{http://www.nrel.gov/grid/solar-power-data.html}} records the solar power production from $137$ PV plants in Alabama State, which are sampled every $10$ minutes in $2016$.
(4)~\emph{Electricity}\footnote{\url{https://archive.ics.uci.edu/ml/datasets/ElectricityLoadDiagrams20112014}} includes the hourly electricity consumption (kWh) records of $321$ clients from $2012$ to $2014$.
(5)~\emph{Exchange-Rate}\footnote{\url{https://github.com/laiguokun/multivariate-time-series-data}} collects the daily exchange rates of $8$ foreign countries from $1990$ to $2016$.
% \emph{Traffic}, \emph{Solar-Energy}, \emph{Electricity} and \emph{Exchange-Rate}~\citep{Lai2018ModelingLA} 
(6)~\textit{PeMS}\footnote{\url{https://pems.dot.ca.gov}} contains four public traffic network datasets (\emph{PEMS03, PEMS04, PEMS07 and PEMS08}) which are respectively constructed from Caltrans Performance Measurement System (PeMS) of four districts in California. The data is aggregated into $5$-minutes windows, resulting in $12$ points per hour and $288$ points per day. 

\subsection{Electricity Transformer Temperature (ETT)}

% \textit{ETT}\footnote{\url{https://github.com/zhouhaoyi/ETDataset}}consists of $2$ year electric power data collected from two separated counties of China, including hourly subsets $\left \{ ETTh1, ETTh2 \right \}$ and quarter-hourly subsets $\left \{ ETTm1 \right \}$. Each data point includes an ``oil temperature'' value and $6$ power load features.

For data pre-processing, we perform zero-mean normalization, i.e., $X^{'} = (X - mean(X))/std(X)$, where $mean(X)$ and $std(X)$ are the mean and the standard deviation of historical time series, respectively.
We use Mean Absolute Errors (MAE)~\citep{Hyndman2006AnotherLA} and Mean Squared Errors (MSE)~\citep{Makridakis1982TheAO} for model comparison. Besides, the train, validation and test sets contain $12$, $4$ and $4$ months data, respectively. 
\begin{equation}
\scriptsize
    MAE = \frac{1}{\tau}\sum_{i =0}^{\tau}|\hat{x}_i -x_i |
\end{equation}
\begin{equation}
\scriptsize
    MSE = \frac{1}{\tau}\sum_{i =0}^{\tau}(\hat{x}_i -x_i)^2
\end{equation}
where $\hat{x}_i$ is the model's prediction, and $x_i$ is the ground-truth. $\tau$ is  the length of the prediction horizon.

% \textbf{Description}:~The data is collected from Electricity Transformers at 2 stations in China, including $\left \{ ETTh1, ETTh2 \right \}$  for 1-hour level and $\left \{ ETTm1 \right \}$ for 15-minutes level. Each data point consists of target value "oil temperature" and 6 power load features. The train/val/test is 12/4/4 months.\\
% \textbf{Normalization}: zero-mean normalization, \\$X = (X - mean(X))/std(X)$.\\
% \textbf{License}: https://github.com/zhouhaoyi/ETDataset.
\subsection{PeMS}

% \textit{PeMS}\footnote{\url{https://pems.dot.ca.gov}} contains four public traffic network datasets (\emph{PEMS03, PEMS04, PEMS07 and PEMS08}) which are respectively constructed from Caltrans Performance Measurement System (PeMS) of four districts in California. The data is aggregated into $5$-minutes windows, resulting in $12$ points per hour and $288$ points per day. 
% We use traffic flow data from the past hour to predict the flow for the next hour. The data is pre-processed using zero-mean normalization as ETT.

Following~\citep{Hyndman2006AnotherLA}, the data is pre-processed using zero-mean normalization and we use Root Mean Squared Errors (RMSE) and Mean Absolute Percentage Errors (MAPE) as evaluation metrics on this dataset. 
\begin{equation}
\scriptsize
    RMSE = \sqrt{\frac{1}{\tau}\sum_{i =0}^{\tau}(\hat{x}_i-x_i)^2},
\end{equation}
\begin{equation}
\scriptsize
    MAPE= \sqrt{\frac{1}{\tau}\sum_{i =0}^{\tau}|(\hat{x}_i-x_i )/x_i|}.
\end{equation}

\subsection{Traffic, Solar-Energy, Electricity and Exchange-Rate}
% \emph{Traffic}\footnote{\url{http://pems.dot.ca.gov}} contains the hourly data describing the road occupancy rates (ranging from $0$ to $1$) that are recorded by the sensors on San Francisco Bay area freeways from $2015$ to $2016$ ($48$ months in total). 
% \emph{Solar-Energy}\footnote{\url{http://www.nrel.gov/grid/solar-power-data.html}} records the solar power production from $137$ PV plants in Alabama State, which are sampled every $10$ minutes in $2016$.
% \emph{Electricity}\footnote{\url{https://archive.ics.uci.edu/ml/datasets/ElectricityLoadDiagrams20112014}} includes the hourly electricity consumption (kWh) records of $321$ clients from $2012$ to $2014$.
% \emph{Exchange-Rate}\footnote{\url{https://github.com/laiguokun/multivariate-time-series-data}} collects the daily exchange rates of $8$ foreign countries from $1990$ to $2016$. %, including Australia, British, Canada, Switzerland, China, Japan, New Zealand and Singapore. 

In our experiments, the length of the look-back window $T$ for these datasets is $168$, and we trained independent models for different length of future horizon (i.e., $\tau=3, 6, 12, 24$).
We use Root Relative Squared Error (RSE) and Empirical Correlation Coefficient (CORR) to evaluate the performance of the TSF models on these datasets following~\citep{Lai2018ModelingLA}, which are calculated as follows:
\begin{equation}
\scriptsize
    RSE = \frac{\sqrt{\sum_{i =0}^{\tau}(\hat{x}_i -x_i)^2}}{\sqrt{\sum_{i =0}^{\tau}(x_i-mean(X))^2}},
\end{equation}
\begin{equation}
\scriptsize
    CORR= \frac{1}{d}\sum_{j=0}^{d} \frac{\sum_{i=0}^{ \tau}(x_{i,j}-mean(X_j))(\hat{x}_{i,j}-mean(\hat{X}_j))}{\sum_{i=0}^{  \tau}(x_{i,j}-mean(X_j))^2(\hat{x}_{i,j}-mean(\hat{X}_j))^2},
\end{equation}
where $X$ and $\hat{X}$ are the ground-truth and model's prediction, respectively. $d$ is the number of variates.

\section{Extra Experimental Results}
\label{sec:extra_results}
In this section, we first add error bars on different forecasting steps T, and also conduct  empirical studies on \emph{ETTh1} and \emph{PEMS} datasets to show the impact of different parameter and operator combinations in \emph{SCI-Block}.

% \subsection{Performance Comparison on Extra Datasets.}
% To further evaluate the performance of the proposed \emph{SCINet}, we also conduct the experiments on $4$ single-step forecasting datasets~\citep{Lai2018ModelingLA}, namely \emph{Traffic}, \emph{Solar-Energy}, \emph{Electricity} and \emph{Exchange-Rate}. The task is defined as follows:

% Given a long time series $\mathbf{X}^*$ and a look-back window of fixed length $T$, at timestamp $t$, the single-step forecasting is to predict the future value $\hat{\mathbf{X}}_{t+\tau:t+\tau} = \{\mathbf{x}_{t+ \tau}\}$. Here, $\tau$ is the length of the forecast horizon, $x_t \in \mathbb{R}^{d}$
% is the value at time step $t$, and $d$ is the number of variates.
\subsection{Error Bars Evaluation}
Since deep models for time series forecasting may be influenced by different random initialization, we report our results with 5 runs on the ETTh1 dataset. From Table~\ref{tab:supp_bar}, we show the standard deviation (Std.) is basically 2\% to 3\% of the mean values, indicating SCINet is robust towards different initialization.

\begin{table}[!ht]
    \centering
    \caption{The error bars of SCINet with 5 runs on the ETTh1 dataset.}
    \begin{tabular}{l|l|l|l|l|l|l|l|l}
    \hline
        T & Metrics & Seed 1 & Seed 2 & Seed 3 & Seed 4 & Seed 5 & Mean & Std. \\ \hline
        24 & MSE & 0.3346 & 0.3381 & 0.3541 & 0.3370 & 0.3370 & 0.3402 & 0.0079 \\ 
        ~ & MAE & 0.3699 & 0.3742 & 0.3826 & 0.3719 & 0.3722 & 0.3742 & 0.0050 \\ \hline
        48 & MSE & 0.4148 & 0.4259 & 0.3899 & 0.3830 & 0.3856 & 0.3998 & 0.0193 \\ 
        ~ & MAE & 0.4370 & 0.4520 & 0.4139 & 0.4108 & 0.4173 & 0.4262 & 0.0177 \\ \hline
        168 & MSE & 0.4490 & 0.5038 & 0.4433 & 0.4493 & 0.4432 & 0.4577 & 0.0259 \\ 
        ~ & MAE & 0.4526 & 0.4985 & 0.4466 & 0.4501 & 0.4476 & 0.4591 & 0.0222 \\ \hline
        336 & MSE & 0.5288 & 0.5935 & 0.5230 & 0.5308 & 0.5373 & 0.5427 & 0.0289 \\ 
        ~ & MAE & 0.5131 & 0.5486 & 0.5114 & 0.5150 & 0.5166 & 0.5209 & 0.0156 \\ \hline
        720 & MSE & 0.5607 & 0.5923 & 0.5855 & 0.5582 & 0.5678 & 0.5729 & 0.0152 \\ 
        ~ & MAE & 0.5469 & 0.5653 & 0.5630 & 0.5418 & 0.5502 & 0.5534 & 0.0103 \\ \hline
    \end{tabular}
    \label{tab:supp_bar}
\end{table}

\subsection{Evaluation on the Impact of $K$ and $L$}~\label{appendix:KL}

We conduct experiments on \emph{ETTh1} dataset (with the multivariate experimental setting) to evaluate the impact of $K$ (number of stacks) and $L$ (number of levels), under various look-back window sizes $T$. The prediction horizon is fixed to be $24$.  

As can be observed from Table~\ref{tab:LK}, when fixing $K=1$,
%~(The best result for $K$ = 1 or $L = 3$ with a certain $T$ is inbold), when $T$ is much larger than the length of the prediction horizon, e.g., $T\!=\!128$ or $192$, 
larger $L$ leads to better prediction accuracy for the cases with larger $T$ ($T\!=\!128$ or $192$).
% However, purely increasing the number of levels $L$ is unnecessary when $T$ is too large.
% For example, for $T\!=\!192$, $L\!=\!4$ performs better than $L\!=\!5$. 
This is because we could further extract essential information from coarser temporal resolutions with deeper levels in the SCINet when $T$ is large. 
%deeper levels will provide more information from different temporal resolutions. Besides, 
As for the number of stacks $K$, when fixing $L=3$, if $T$ is small (e.g. $T\!=\!24$ or $48$), we find that increasing $K$ would improve prediction accuracy. This is because, under such circumstances, the information extracted from a single SCINet is insufficient. By stacking more SCINets, we effectively increase the representation learning capability of the model, which facilitates extracting more robust temporal relations for the forecasting task. 
However, 
% when $T$ is large, it is not necessary to use a large $K$,
when $T$ is large~(e.g., 192), a shallow stack can already well capture the temporal dependencies for the time series. Under such circumstances, using deeper stacks may suffer from overfitting issues with the increase of parameters, which degrades the performance in the inference stage.

\begin{table}[h]
\centering
\footnotesize
\caption{The impact of $L$ and $K$ on MSE.}
\begin{threeparttable}
\begin{tabular}{c|c|c|c|c|c|c}

\hline
\multirow{2}{*}{\begin{tabular}[c]{@{}c@{}}Number of \\ Levels \& Stacks\end{tabular}} & Horizon & \multicolumn{5}{c}{24}                                                                 \\ \cline{2-7} 
                                                                                       & $T$   & 24              & 48              & 96              & 128             & 192             \\ \hline
\multirow{4}{*}{\begin{tabular}[c]{@{}c@{}}Level $L$\\ ($K$ =1)\end{tabular}}            & 2       & 0.411          & 0.348          & 0.347          & 0.334          & 0.384          \\ \cline{2-7} 
                                                                                       & 3       & \textbf{0.405} & \textbf{0.346} & \textbf{0.316}          & 0.418          & 0.330          \\ \cline{2-7} 
                                                                                       & 4       & -               & 0.360          & 0.340 & 0.331          & \textbf{0.325} \\ \cline{2-7} 
                                                                                       & 5       & -               & -               & 0.354          & \textbf{0.323} & 0.356          \\ \hline \hline
\multirow{4}{*}{\begin{tabular}[c]{@{}c@{}}Stack $K$\\ ($L$ = 3)\end{tabular}}           & 1        & 0.405 & 0.346 & \textbf{0.316}          & 0.418          & \textbf{0.330} \\ \cline{2-7} 
                                                                                       & 2       & 0.423          & 0.344          & 0.344 & \textbf{0.339} & 0.375          \\ \cline{2-7} 
                                                                                       & 3       & \textbf{0.374}          & \textbf{0.341} & 0.345          & 0.353          & 0.363          \\ \cline{2-7} 
                                                                                       & 4       & 0.390          & 0.342          & 0.335          & 0.356          & 0.388          \\ \hline

\end{tabular}
\begin{tablenotes} %添加此处
\tiny
		\item  - Dash denotes the input cannot be further splitted. 
	%	\item  - Table shows the MSE results. 
     \end{tablenotes} %添加此处

\end{threeparttable} %添加此处
\label{tab:LK}
\end{table}

From Table~\ref{tab:LK}, we can observe a clear trade-off between $L$ and $K$. Moreover, the performance variation under different $T$ also indicates the importance of the look-back window selection for forecasting tasks. While $T$ is typically pre-determined based on domain knowledge about the time series data, based on our empirical study, $ L\leq 5$ and $K \leq 3$ are usually sufficient and tuning these hyperparameters does not incur much effort.

\subsection{Empirical Study on Operator Selection} \label{appendix:B3}
In interactive-learning equation,
\begin{equation}\label{eq:inter_2}
\small
    \mathbf{F}^{'}_{odd} = \mathbf{F}^{s}_{odd} \pm  \rho( \mathbf{F}^{s}_{even}),~~~
    \mathbf{F}^{'}_{even} = \mathbf{F}^{s}_{even} \pm   \eta(\mathbf{F}^{s}_{odd}).
\end{equation}\textbf{}

the operators can be either "addition" or "subtraction". Although the model can learn the operation adaptively during training, the parameter initialization would affect the final performance. As shown in the following table %Table~\ref{tab:operator}, 
the impact of operator settings is minor.

\begin{table}[h]
\footnotesize
\caption{The impact of different operators}
\centering
% \vspace{-5pt}
\begin{tabular}{c|c|c|c|c}
\hline
\multirow{2}{*}{Operators} & PEMS03            & PEMS04            & PEMS07           & PEMS08           \\ \cline{2-5} 
                           & \multicolumn{4}{c}{MAE}                                          \\ \hline
+, +                       & 15.08 & 19.27 & 21.69 &  \textbf{15.72} \\ \hline
-, -                       & \textbf{15.06} &  \textbf{19.21} &  \textbf{21.63} & 15.78 \\ \hline
+, -                       & 15.09 & 19.31 & 21.77 & 15.84 \\ \hline
-, +                       & 15.30 & 19.32 & 21.72 & 15.79 \\ \hline
\end{tabular}
\label{tab:operator}
\end{table}

\section{Reproducibility}
\label{appendix:reprod}
Our code is implemented with PyTorch. All the experiments are conducted on an Nvidia Tesla V100 SXM2 GPU (32GB memory), which is sufficient for all our experiments. 
%The detailed hyperparameters are introduced in the following.

\begin{figure}[h]
\centering
\includegraphics[scale=0.75]{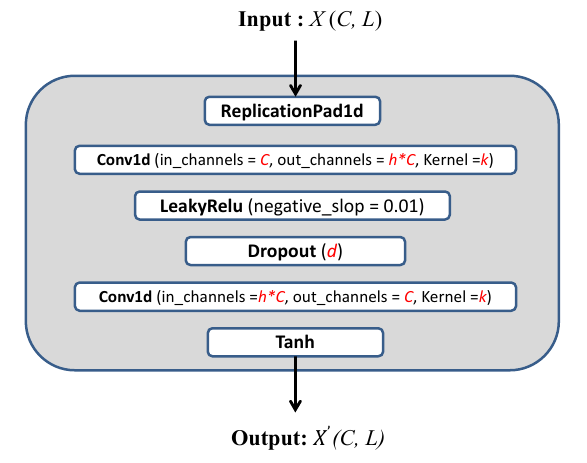}
\caption{The structure of $\phi$, $\rho$, $\psi$, and $\eta$.}
\label{fig:PU}
\end{figure}

\textbf{Structure of the network modules $\phi$, $\rho$, $\psi$, and $\eta$ in SCI-Block:}~As shown in Fig.~\ref{fig:PU}, $\phi$, $\rho$, $\psi$, and $\eta$ use the same network architecture.
First, the replication padding is used to keep the border shrunk caused by the convolution operation. Then, a 1d convolutional layer with kernel size $k$ is applied to extend the input channel $C$ to $h$*$C$ and followed with LeakyRelu and Dropout. $h$ means a scale of the hidden size. Next, the second 1d convolutional layer with kernel size $k$ is to recover the channel $h$*$C$ to the input channel $C$. The stride of all the convolutions is  $1$. We use a LeakyRelu activation after the first convolutional layer because of its sparsity properties and a reduced likelihood of vanishing gradient. We apply a Tanh activation after the second convolutional layer since it can keep both positive and negative features into [-1, 1].

\textbf{Loss Function}

To enhance the performance in single-step (short-term time series forecasting Sec.~\ref{sec:comparison_sota}) forecasting, we revise the loss function of the last SCINet in the stacked SCINet with $K$($K \geq 1$). The loss function contains two parts:
% To train a stacked SCINet with $K$ ($K\geq 1$) SCINets, the loss of the $k$-th intermediate prediction is calculated as the L1 loss between the output of the $k$-th SCINet and the ground-truth horizontal window to be predicted:
\begin{equation} \label{eq:mid_loss}
            % \mathcal{L}_{k} = \red{\frac{1}{\tau}} \left \| \hat{\mathbf{X}}^{k+1}  - \hat{\mathbf{X}}  \right \|, ~~~1\leq k<K .
            \mathcal{L}_{k} =  \frac{1}{\tau}\sum_{i=0}^{\tau}\left \| \hat{\mathbf{x}}^{k}_{i} - \mathbf{x}_{i} \right \|,~~~k\neq K .
\end{equation}
% where $\mathbf{X}^{k+1}$ is the output of the $k$-th SCINet, and $\hat{\mathbf{X}}$ is the horizontal window to be predicted.
For the last stack $K$, we introduce a balancing parameter $\lambda \in (0,1)$ for the value of the last time-step\footnote{This is slightly different from other practice for single-step forecasting~\citep{Lai2018ModelingLA}, because we choose to use all the available values in the prediction window as supervision signal.}:
\begin{equation}\label{eq:last_loss}
             \mathcal{L}_{K} = \frac{1}{\tau-1}\sum_{i=0}^{\tau-1}\left \| \hat{\mathbf{x}}^{K}_{i} - \mathbf{x}_{i} \right \| + \lambda \left \| \hat{\mathbf{x}}^{K}_{\tau} - \mathbf{x}_{\tau} \right \|.
\end{equation}

Therefore, the total loss of the stacked SCINet can be written as:
\begin{equation}
     \mathcal{L} = \sum_{k=1}^{K-1}\mathcal{L}_{k} +  \mathcal{L}_{K}.
\end{equation}

% \textbf{Experimental details:}
\textbf{Training details:}
% For \emph{ETT}, we follow the same training strategy as \cite{Zhou2020InformerBE}, and all the experiments are conducted over five random train/val shifting selection along time. The results are averaged over the five runs. 
% % For \emph{Traffic, Solar-Energy, Electricity, Exchange-Rate} and \emph{PeMs}, 
For all datasets, we fix the random seed to be 4321, and train the model for 150 epochs at most. The reported results on the test set are based on the model that achieves the best performance on the validation set.

\textbf{Hyper-parameter tuning:}
We conduct a grid search over all the essential hyper-parameters on the held-out validation set of the datasets. The detailed hyper-parameter configurations of \emph{ETT} are shown in Table~\ref{tab:hetth} \footnote{The results on ETTh2 and ETTm1 datasets can be referred to: \url{https://github.com/cure-lab/SCINet} }. Besides, the parameters of the four datasets in \emph{PeMS} are presented in  Table\ref{tab:pems}. The \emph{Traffic, Solar-Energy, Electricity and Exchange-rate} are shown in Table~\ref{tab:hsolar}.  
% The performance of weighted loss varies in different data. 
Notably, we only apply the weighted loss to the \textit{Solar} and \textit{Exchange-rate} data since they show less auto-correlation~\cite{Lai2018ModelingLA}, which indicates the temporal correlation of the distant time-stamp cannot be well modelled by a general L1 loss. 
Moreover, to build a non-causal TCN\footnote{https://github.com/locuslab/TCN/issues/45} in the paper, we only need to remove the \emph{chomps} in the code and make the padding equal to the dilation.

% For other re-implementation baselines, e.g., LSTM in \emph{PeMS} dataset, the hidden state is chosen from $\left \{ 32, 64, 128, 256 \right \}$.
% For TCN, in order to cover different length of the look-back window, the number of layers is chosen from $\left \{ 3, 4, 5 \right \}$ and the dimension of the hidden state is chosen from $\left \{ 32, 64, 128, 256 \right \}$. 

\newpage

\begin{table*}[h]
\centering
\caption{The hyperparameters in ETT datasets (Multivariate)}
\resizebox{\textwidth}{!}
{
\begin{tabular}{c|c|c|c|c|c|c|c|c|c|c|c|c|c|c|c|c}
\hline
\multicolumn{2}{c|}{Model configurations}          & \multicolumn{5}{c|}{ETTh1}       & \multicolumn{5}{c|}{ETTh2}       & \multicolumn{5}{c}{ETTm1}       \\ \hline
\multirow{4}{*}{Hyperparameter} & Horizon          & 24   & 48   & 168  & 336  & 720  & 24   & 48   & 168  & 336  & 720  & 24   & 48   & 96   & 288  & 672  \\ \cline{2-17} 
                                & Look-back window & 48   & 96   & 336  & 336  & 736  & 48   & 96   & 336  & 336  & 736  & 48   & 96   & 384  & 672  & 672  \\ \cline{2-17} 
                                & Batch size       & 8    & 16   & 32   & 512  & 256  & 16   & 4    & 16   & 128  & 128  & 32   & 16   & 32   & 32   & 32   \\ \cline{2-17} 
                                & Learning rate    & 3e-3 & 9e-3 & 5e-4 & 1e-4 & 5e-5 & 7e-3 & 7e-3 & 5e-5 & 5e-5 & 1e-5 & 5e-3 & 1e-3 & 5e-5 & 1e-5 & 1e-5 \\ \hline
\multirow{3}{*}{SCI Block}      & h                & 4    & 4    & 4    & 1    & 1    & 8    & 4    & 0.5  & 1    & 4    & 4    & 4    & 0.5  & 4    & 4    \\ \cline{2-17} 
                                & k                & 5    & 5    & 5    & 5    & 5    & 5    & 5    & 5    & 5    & 5    & 5    & 5    & 5    & 5    & 5    \\ \cline{2-17}
                                & Dropout          & 0.5  & 0.25 & 0.5  & 0.5  & 0.5  & 0.25 & 0.5  & 0.5  & 0.5  & 0.5  & 0.5  & 0.5  & 0.5  & 0.5  & 0.5    \\ \hline
SCINet                          & L (level)        & 3    & 3    & 3    & 4    & 5    & 3    & 4    & 4    & 4    & 5    & 3    & 4    & 4    & 5    & 5    \\ \hline
Stacked SCINet                  & K (stack)        & 1    & 1    & 1    & 1    & 1    & 1    & 1    & 1    & 1    & 1    & 1    & 2    & 2    & 1    & 2    \\ \hline
\end{tabular}}
\label{tab:hetth}
\vspace{-8pt}
\end{table*}

\begin{table*}[h]
\centering
\caption{The hyperparameters in Traffic, Solar-energy, Electricity and Exchange-rate datasets}
\resizebox{\textwidth}{!}
{
% \begin{threeparttable}
\begin{tabular}{ccclllclllclllclll}
\hline
\multicolumn{2}{c|}{Model configurations}                                                                        & \multicolumn{4}{c|}{Solar}                                                                        & \multicolumn{4}{c|}{Electricity}                                                                          & \multicolumn{4}{c|}{Traffic}                                                                    & \multicolumn{4}{c}{Exc-Rate}                                                                      \\ \hline
\multicolumn{1}{c|}{\multirow{4}{*}{Hyperparameter}} & \multicolumn{1}{c|}{Horizon}                              & \multicolumn{1}{c|}{3} & \multicolumn{1}{c|}{6} & \multicolumn{1}{c|}{12} & \multicolumn{1}{c|}{24} & \multicolumn{1}{c|}{3} & \multicolumn{1}{c|}{6} & \multicolumn{1}{c|}{12} & \multicolumn{1}{c|}{24} & \multicolumn{1}{c|}{3} & \multicolumn{1}{c|}{6} & \multicolumn{1}{c|}{12} & \multicolumn{1}{c|}{24} & \multicolumn{1}{c|}{3} & \multicolumn{1}{c|}{6} & \multicolumn{1}{c|}{12} & \multicolumn{1}{c}{24} \\ \cline{2-18} 
\multicolumn{1}{c|}{}                                & \multicolumn{1}{c|}{Look-back window}                     & \multicolumn{4}{c|}{160}   & \multicolumn{12}{c}{168}                                                                                                                                                                                                                                                                                                                                                                                            \\ \cline{2-18} 
\multicolumn{1}{c|}{}                                & \multicolumn{1}{c|}{Batch size}                           & \multicolumn{1}{c|}{256}   & \multicolumn{1}{c|}{256}& \multicolumn{1}{c|}{1024}& \multicolumn{1}{c|}{256}                                                                          & \multicolumn{4}{c|}{32}                                                                           & \multicolumn{4}{c|}{16}                                                                             & \multicolumn{4}{c}{4}                                                                             \\ \cline{2-18} 
\multicolumn{1}{c|}{}                                & \multicolumn{1}{c|}{Learning rate}                        & \multicolumn{4}{c|}{1e-4}                                                                           & \multicolumn{4}{c|}{9e-3}                                                                           & \multicolumn{4}{c|}{5e-4}                                                                           & \multicolumn{3}{c|}{5e-3}    & \multicolumn{1}{c}{7e-3}                                                                      \\ \hline
\multicolumn{1}{c|}{\multirow{3}{*}{SCI Block}}      & \multicolumn{1}{c|}{h}                                    & \multicolumn{1}{c|}{1}   & \multicolumn{1}{c|}{0.5}& \multicolumn{1}{c|}{2}& \multicolumn{1}{c|}{1}                                                                           & \multicolumn{4}{c|}{8}                                                                              & \multicolumn{1}{c|}{1}   & \multicolumn{1}{c|}{2}& \multicolumn{1}{c|}{0.5}& \multicolumn{1}{c|}{2}                                                                           & \multicolumn{4}{c}{0.125}                                                                         \\ \cline{2-18} 
\multicolumn{1}{c|}{}                                & \multicolumn{1}{c|}{k}                                    & \multicolumn{4}{c|}{5}                                                                              & \multicolumn{4}{c|}{5}                                                                              & \multicolumn{4}{c|}{5}                                                                              & \multicolumn{4}{c}{5}                                                                             \\ \cline{2-18}
\multicolumn{1}{c|}{}                                & \multicolumn{1}{c|}{Dropout}                                    & \multicolumn{4}{c|}{0.25}                                                                              & \multicolumn{4}{c|}{0}                                                                              & \multicolumn{1}{c|}{0.5}         & \multicolumn{1}{c|}{0.25} & \multicolumn{1}{c|}{0.25} & \multicolumn{1}{c|}{0.5}                                                                     & \multicolumn{4}{c}{0.5}                                                                             \\ \hline
\multicolumn{1}{c|}{SCINet}                          & \multicolumn{1}{c|}{L (level)}                            & \multicolumn{4}{c|}{4}                                                                              & \multicolumn{4}{c|}{3}                                                                              & \multicolumn{3}{c|}{3}  & \multicolumn{1}{c|}{2}                                                                            & \multicolumn{4}{c}{3}                                                                             \\ \hline
\multicolumn{1}{c|}{\multirow{2}{*}{Stacked SCINet}} & \multicolumn{1}{c|}{K (stack)}                            & \multicolumn{3}{c|}{2}  & \multicolumn{1}{c|}{1}                                                                            & \multicolumn{4}{c|}{2}                                                                              & \multicolumn{1}{c|}{2}& \multicolumn{1}{c|}{1}& \multicolumn{1}{c|}{2}& \multicolumn{1}{c|}{2}                                                                              & \multicolumn{4}{c}{1}                                                                             \\ \cline{2-18} 
\multicolumn{1}{c|}{}                                & \multicolumn{1}{c|}{Loss weight ($\lambda)$} & \multicolumn{4}{c|}{0.5}                                                          & \multicolumn{4}{c|}{$\times$}                                                                            & \multicolumn{4}{c|}{$\times$}                                                          & \multicolumn{4}{c}{0.5}                                                                           \\ \hline
\multicolumn{1}{l}{}                                 & \multicolumn{1}{l}{}                                      & \multicolumn{1}{l}{}   &                        &                         &                         & \multicolumn{1}{l}{}   &                        &                         &                         & \multicolumn{1}{l}{}   &                        &                         &                         & \multicolumn{1}{l}{}   &                        &                         &                        \\
\multicolumn{1}{l}{}                                 & \multicolumn{1}{l}{}                                      & \multicolumn{1}{l}{}   &                        &                         &                         & \multicolumn{1}{l}{}   &                        &                         &                         & \multicolumn{1}{l}{}   &                        &                         &                         & \multicolumn{1}{l}{}   &                        &                         &                        \\
\multicolumn{1}{l}{}                                 & \multicolumn{1}{l}{}                                      & \multicolumn{1}{l}{}   &                        &                         &                         & \multicolumn{1}{l}{}   &                        &                         &                         & \multicolumn{1}{l}{}   &                        &                         &                         & \multicolumn{1}{l}{}   &                        &                         &                       
\end{tabular}}
% \begin{tablenotes} %添加此处
% 		\item   $\times$ denotes the dataset does not combine with weighted loss.
%  \end{tablenotes} %添加此处
% \end{threeparttable}} %添加此处
\label{tab:hsolar}
\vspace{-24pt}
\end{table*}

% Optionally include extra information (complete proofs, additional experiments and plots) in the appendix.
% This section will often be part of the supplemental material.

\begin{table*}[h]
\centering
\caption{The hyperparameters in PeMS datasets}
\resizebox{0.7\textwidth}{!}
{
\begin{tabular}{c|c|c|c|c|c}
\hline
\multicolumn{2}{c|}{Model configurations}          & PEMS03       & PEMS04       & PEMS07 & PEMS08       \\ \hline
\multirow{4}{*}{Hyperparameter} & Horizon          & \multicolumn{4}{c}{12} \\ \cline{2-6} 
                                & Look-back window & \multicolumn{4}{c}{12}  \\ \cline{2-6} 
                                & Batch size       & \multicolumn{4}{c}{8}  \\ \cline{2-6}
                                & Learning rate    & \multicolumn{4}{c}{1e-3}  \\ \hline
\multirow{3}{*}{SCI Block}      & h                & 0.0625    & 0.0625    & 0.03125    & 1   \\ \cline{2-6} 
                                & k                & \multicolumn{4}{c}{5}   \\ \cline{2-6}
                                & Dropout       & 0.25 & 0 & 0.25 & 0.5 \\ \hline
SCINet                          & L (level)        & \multicolumn{4}{c}{2}   \\ \hline
Stacked SCINet                  & K (stack)        & \multicolumn{4}{c}{1}    \\ \hline
\end{tabular}}
\label{tab:pems}
\vspace{-8pt}
\end{table*}

\end{document}